\pdfoutput=1

\documentclass[11pt]{article}

\usepackage{acl}

\usepackage{times}
\usepackage{latexsym}

\usepackage[disable]{todonotes}
\usepackage[T1]{fontenc}
\usepackage[T5]{fontenc}
\usepackage[utf8]{inputenc}

\usepackage{microtype}

\usepackage{makecell}

\usepackage{pifont}

\usepackage{color}

\newcommand{\forcamera}[1]{}
\usepackage{float}
\usepackage{makecell}

\usepackage{pgfplots}
\pgfplotsset{compat=1.16}
\usepackage{tikz}
\usepackage{amsmath}

\usepackage{flafter} %

\newcommand{\caancora}[0]{\textsf{ca\_ancora}}%
\newcommand{\cspcedt}[0]{\textsf{cs\_pcedt}}%
\newcommand{\cspdt}[0]{\textsf{cs\_pdt}}%
\newcommand{\cuproiel}[0]{\textsf{cu\_proiel}}%
\newcommand{\deparcorfull}[0]{\textsf{de\_parcorfull}}%
\newcommand{\depotsdamcc}[0]{\textsf{de\_potsdam}}%
\newcommand{\engum}[0]{\textsf{en\_gum}}%
\newcommand{\enlitbank}[0]{\textsf{en\_litbank}}%
\newcommand{\enparcorfull}[0]{\textsf{en\_parcorfull}}%
\newcommand{\esancora}[0]{\textsf{es\_ancora}}%
\newcommand{\francor}[0]{\textsf{fr\_ancor}}%
\newcommand{\frdemocrat}[0]{\textsf{fr\_democrat}}%
\newcommand{\grcproiel}[0]{\textsf{grc\_proiel}}%
\newcommand{\hboptnk}[0]{\textsf{hbo\_ptnk}}%
\newcommand{\hihdtb}[0]{\textsf{hi\_hdtb}}%
\newcommand{\huszegedkoref}[0]{\textsf{hu\_szeged}}%
\newcommand{\hukorkor}[0]{\textsf{hu\_korkor}}%
\newcommand{\koecmt}[0]{\textsf{ko\_ecmt}}%
\newcommand{\ltlcc}[0]{\textsf{lt\_lcc}}%
\newcommand{\nobokmaalnarc}[0]{\textsf{no\_bokmaalnarc}}%
\newcommand{\nonynorsknarc}[0]{\textsf{no\_nynorsk\-narc}}%
\newcommand{\plpcc}[0]{\textsf{pl\_pcc}}%
\newcommand{\rurucor}[0]{\textsf{ru\_rucor}}%
\newcommand{\tritcc}[0]{\textsf{tr\_itcc}}%

\newcommand{\sys}[1]{\textsc{#1}}
\newcommand{\baseline}[0]{\sys{Baseline}}
\newcommand{\baselinegz}[0]{\sys{Baseline-GZ}}

\def\MC#1#2{\multicolumn{#1}{c}{#2}}

\title{Findings of the Fourth Shared Task on Multilingual Coreference Resolution:
Can LLMs Dethrone Traditional Approaches?}

\author{
Michal Novák$^1$,
Miloslav Konopík$^2$,
Anna Nedoluzhko$^1$,
Martin Popel$^1$, \\
\textbf{Ondřej Pražák$^2$,
Jakub Sido$^2$,
Milan Straka$^1$,
Zdeněk Žabokrtský$^1$,
Daniel Zeman$^1$} \\[2mm]
$^1$ Charles University, Faculty of Mathematics and Physics, \\ Institute of Formal and Applied Linguistics, Prague, Czechia \\
\texttt{\{mnovak,nedoluzko,popel,straka,zabokrtsky,zeman\}@ufal.mff.cuni.cz}\\[2mm]
$^2$
University of West Bohemia, Faculty of Applied Sciences, \\ Department of Computer Science and Engineering, Pilsen, Czechia \\
\texttt{\{konopik,ondfa,sidoj\}@kiv.zcu.cz}\\[2mm]
}

\usepackage{booktabs}
\newcommand{\ndatasets}[0]{22}
\newcommand{\nlanguages}[0]{17} %
\newcommand{\nsystems}[0]{nine} %

\usepackage{fancyhdr}
\fancypagestyle{officialbibref}{\fancyhf{}\fancyheadoffset[L,R]{50pt}\fancyhead[C]{This paper was published at \textbf{CODI-CRAC 2025} -- please cite the published version {\small\url{https://aclanthology.org/2025.crac-1.9}}.}\fancyfoot[C]{\normalsize\thepage}\addtolength{\topmargin}{-30pt}\addtolength{\headsep}{30pt}}
\begin{document}
\thispagestyle{officialbibref}
\pagenumbering{arabic}\pagestyle{plain}
\maketitle

\begin{abstract}

The paper presents an overview of the fourth edition of the Shared Task on
Multilingual Coreference Resolution, organized as part of the CODI-CRAC 2025
workshop. 
As in the previous editions, participants were challenged to develop systems that identify mentions and cluster them according to identity coreference.

A key innovation of this year's task was the introduction of a dedicated Large Language Model (LLM) track, featuring a simplified plaintext format designed to be more suitable for LLMs than the original CoNLL-U representation.

The task also expanded its coverage with three new datasets in two additional languages, using version 1.3 of CorefUD -- a harmonized multilingual collection of \ndatasets{} datasets in \nlanguages{} languages.

In total, \nsystems{} systems participated, including four LLM-based approaches (two fine-tuned and two using few-shot adaptation).
While traditional systems still kept the lead,
LLMs showed clear potential, suggesting they may soon challenge established approaches in future editions.

\end{abstract}

\section{Introduction}

\begin{figure*}
    \centering
    \includegraphics[width=\linewidth]{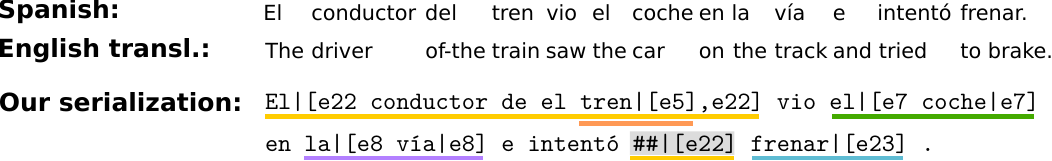}
    \caption{Our plaintext serialization of a Spanish example sentence from \esancora{}. For clarity, mention spans are highlighted by colored underlining, where two coreferential entities share the same color. A zero mention labeled on an empty node is greyed. Note that multi-word tokens are split in the plaintext format into syntactic words (e.g., the Spanish \emph{``del''} appears as \emph{``de el''}); this conversion error was identified after the data release.}
    \label{fig:plaintext-example}
\end{figure*}

Coreference is the phenomenon where multiple expressions in a text refer to the same real-world entity. For example: \emph{``Beethoven was a revolutionary artist. The German composer changed the course of music, and he continues to inspire musicians today.''} Here, \emph{``Beethoven''}, \emph{``the German composer''}, and \emph{``he''} all point to the same individual.
The computational task of coreference resolution is to automatically identify such links between mentions and group them into clusters that represent entities. In the multilingual setting, the task is the same, but complicated by the diversity of languages and their grammatical and discourse conventions.

In this article, we present the overall setup and results of the fourth edition of the shared task in multilingual coreference resolution. For descriptions of previous editions, as well as references to the roots and predecessors of the series, see \citet{oursharedtask2024}.

This year's edition uses an improved and expanded collection of coreference data, CorefUD~1.3 \cite{corefud1.3}, currently spanning 17 languages from a few typologically different families.  However, the most important novelty in this edition is the introduction of the Large Language Model (LLM) track. Although non-LLM models were still welcome, a dedicated LLM Track was introduced to highlight and explore the capabilities of LLM-based approaches. Hence, to accommodate different modeling strategies and study their effects, we defined two shared-task tracks:
 \begin{itemize}
     \item \emph{LLM Track:} Focused on solutions that primarily rely on LLMs for coreference resolution. Allowed strategies include fine-tuning LLMs, using in-context learning, designing effective prompts, utilizing constrained decoding strategies, and building more complex agentic systems.
    \item \emph{Unconstrained Track:} Open to all other approaches, including non-LLM models and hybrid systems. This track allows the use of additional pre-existing coreference systems, external tools, and extensive model modifications.
\end{itemize}

A major trend in NLP is the shift from traditional task-specific models to LLMs, which can address a wide range of tasks with little fine-tuning and are comparatively easy to deploy. This unification brings greater efficiency, flexibility, and scalability, but also raises challenges such as bias, computational cost, and privacy concerns. At the same time, LLMs have shown strong performance on tasks that require understanding of textual context and relations, including question answering, summarization, and commonsense reasoning.
\todo{Mišo: cite}

A category of benchmarks that are commonly used to test these coreference-related capabilities are derivations of the Winograd Schema Challenge \citep{levesque.etal12}, for instance KnowRef \citep{emami.etal19}, WinoGrande \citep{sakaguchi.etal21}, and recently WinoWhat \citep{gevers.etal25}.
However, these benchmarks represent an overly narrow view of coreference resolution.
They primarily focus on commonsense reasoning through carefully crafted disambiguation scenarios, while real-world coreference resolution involves a much broader spectrum of phenomena.

Previous works on using LLMs for coreference resolution show that they struggle with this task and are not able to outperform systems specifically tailored for coreference resolution \citep{le_are_2023,vadasz2023resolving,hicke_lions_2024,gan_assessing_2024,saputa_polish_2024}.
One of the reasons may be that the data used to model and test the task is very heterogeneous due to practical difficulties in clearly and precisely defining the elements that coreference relations work with, specifically the scope of mentions, the degree of zero reconstruction, and the typology of coreference and anaphoric relations.

Still, the progress in LLMs is so rapid that it seems just a matter of time before these LLM-based systems will dominate also in this task.
We see the LLM track of this shared task as an opportunity to test this hypothesis and encourage development in this field, providing a platform for researchers to explore the current boundaries and future potential of LLM-based coreference resolution.

The step towards LLMs does not represent only a technological change -- it often requires rethinking how we approach a particular task. Structured (possibly pipelined) solutions are typically abandoned and replaced by processing ``flat'' sequences of (sub)words. In the particular case of this shared task, we replace the relatively richly structured CoNLL-U format in which the encoding of coreference relations is stored in the CorefUD collection with an encoding of coreference that could be added directly into plain text.

Naturally, there are many possible ways to insert coreference markup into text, and prior work on LLMs for coreference has each used its own prompt and format. So far, no widely accepted best practices have emerged for encoding or prompting coreference in plain text. We implement our own conversion from CorefUD into a plaintext serialization (example in Figure~\ref{fig:plaintext-example}), but acknowledge that our design choices may limit applicability and that further optimization could improve LLM performance.

The remainder of the paper is structured as follows.  Section~\ref{sec:data}
discusses the changes in the data compared to the previous (third)
edition of the shared task. Section~\ref{sec:evaluation} outlines the evaluation metrics used in the task, including both the primary and supplementary scores.
Section~\ref{sec:systems} details all participating
systems, both in the LLM track and in the Unconstrained track. Section~\ref{sec:results} presents a summary of the results and discusses some differences between the performance of LLM and Unconstrained systems. Section~\ref{sec:conclusions} provides the conclusion.

\section{Datasets}
\label{sec:data}

\begin{table*}[!htb]
  \begin{center}
    \resizebox{\textwidth}{!}{
    \begin{tabular}{@{}l rrrr rrrr rrrr @{}}\toprule
                                  & \MC{4}{total number of}              & \MC{4}{entities}                 & \MC{4}{mentions}                  \\\cmidrule(lr){2-5}\cmidrule(lr){6-9}\cmidrule(lr){10-13}
    document                      &        &         &         &         &  total & per 1k & \MC{2}{length} &   total & per 1k & \MC{2}{length} \\\cmidrule(lr){8-9}\cmidrule(lr){12-13}
                                  &   docs &   sents &   words & empty n.&  count &  words &    max &  avg. &   count &  words &    max &  avg. \\\midrule
    Ancient\_Greek-PROIEL         &     19 &   6,475 &  64,111 &   6,283 &   3,215 &     50 &    332 &   6.6 &  21,354 &    333 &     52 &   1.7 \\
    Ancient\_Hebrew-PTNK          &     40 &   1,161 &  28,485 &       0 &     870 &     31 &    102 &   7.2 &   6,247 &    219 &     22 &   1.5 \\
    Catalan-AnCora                &  1,298 &  13,613 & 429,313 &   6,377 &  17,558 &     41 &    101 &   3.6 &  62,417 &    145 &    141 &   4.8 \\
    Czech-PCEDT                   &  2,312 &  49,208 & 1,155,755 &  35,654 &  49,225 &     43 &    236 &   3.4 & 168,055 &    145 &     79 &   3.6 \\
    Czech-PDT                     &  3,165 &  49,419 & 834,707 &  21,092 &  46,460 &     56 &    173 &   3.3 & 154,437 &    185 &     99 &   3.1 \\
    English-GUM                   &    237 &  13,263 & 233,926 &     119 &   9,200 &     39 &    131 &   4.4 &  40,656 &    174 &     95 &   2.6 \\
    English-LitBank               &    100 &   8,560 & 210,530 &       0 &   2,164 &     10 &    261 &  10.8 &  23,340 &    111 &    129 &   1.6 \\
    English-ParCorFull            &     19 &     543 &  10,798 &       0 &     188 &     17 &     38 &   4.4 &     835 &     77 &     37 &   2.1 \\
    French-ANCOR                  &    455 &  31,761 & 454,577 &       0 &  13,204 &     29 &    103 &   4.3 &  56,459 &    124 &     17 &   1.9 \\
    French-Democrat               &    126 &  13,057 & 284,883 &       0 &   7,162 &     25 &    895 &   6.5 &  46,487 &    163 &     71 &   1.7 \\
    German-ParCorFull             &     19 &     543 &  10,602 &       0 &     243 &     23 &     43 &   3.7 &     896 &     85 &     30 &   2.0 \\
    German-PotsdamCC              &    176 &   2,238 &  33,222 &       0 &     880 &     26 &     15 &   2.9 &   2,519 &     76 &     34 &   2.6 \\
    Hindi-HDTB                    &    271 &   3,479 &  76,282 &       0 &   3,148 &     41 &     36 &   3.8 &  12,082 &    158 &     43 &   1.8 \\
    Hungarian-KorKor              &     94 &   1,351 &  24,568 &   1,569 &   1,122 &     46 &     41 &   3.6 &   4,091 &    167 &     42 &   2.2 \\
    Hungarian-SzegedKoref         &    400 &   8,820 & 123,968 &   4,857 &   4,769 &     38 &     36 &   3.2 &  15,165 &    122 &     36 &   1.6 \\
    Korean-ECMT                   &  1,470 &  30,784 & 482,986 &       0 &  16,536 &     34 &     55 &   3.4 &  56,538 &    117 &     12 &   1.3 \\
    Lithuanian-LCC                &    100 &   1,714 &  37,014 &       0 &   1,087 &     29 &     23 &   4.0 &   4,337 &    117 &     19 &   1.5 \\
    Norwegian-BokmaalNARC         &    346 &  15,742 & 245,515 &       0 &   5,658 &     23 &    298 &   4.7 &  26,611 &    108 &     51 &   1.9 \\
    Norwegian-NynorskNARC         &    394 &  12,481 & 206,660 &       0 &   5,079 &     25 &     84 &   4.3 &  21,847 &    106 &     57 &   2.1 \\
    Old\_Church\_Slavonic-PROIEL  &     26 &   6,832 &  61,759 &   6,289 &   3,396 &     55 &    134 &   6.5 &  22,116 &    358 &     52 &   1.5 \\
    Polish-PCC                    &  1,828 &  35,874 & 538,885 &  18,615 &  22,143 &     41 &    135 &   3.7 &  82,706 &    153 &    108 &   1.9 \\
    Russian-RuCor                 &    181 &   9,035 & 156,636 &       0 &   3,515 &     22 &    141 &   4.6 &  16,193 &    103 &     18 &   1.7 \\
    Spanish-AnCora                &  1,356 &  14,159 & 458,418 &   8,112 &  19,445 &     42 &    110 &   3.6 &  70,663 &    154 &    101 &   4.8 \\
    Turkish-ITCC                  &     24 &   4,732 &  55,358 &  11,584 &   4,019 &     73 &    369 &   5.4 &  21,569 &    390 &     31 &   1.1 \\
    \bottomrule
    \end{tabular}
    }
    \caption{CorefUD~1.3 data sizes in terms of the total number of documents, sentences,
      words (i.e.\ non-empty nodes), empty nodes (empty words),
      coreference entities
      (total count, relative count per 1000 words, average and maximal length in number of mentions)
      and coreference mentions
      (total count, relative count per 1000 words, average and maximal length in number of words).
      All the counts are excluding singletons and for the concatenation of train+dev+test.
      Train/dev/test splits of these datasets roughly follow the 8/1/1 ratio.
      However, note that for the shared task we used reduced versions of dev and test: mini-dev and mini-test, respectively.
      }
    \label{tab:sizes}
  \end{center}
\end{table*}

As in previous years, the shared task takes training and evaluation
data from the public part of the CorefUD collection
\cite{corefud2022lrec},\footnote{\url{https://ufal.mff.cuni.cz/corefud}} now
in its latest release
(1.3).\footnote{\url{http://hdl.handle.net/11234/1-5896}} The public edition
of CorefUD~1.3 consists of 24 datasets%
\footnote{For the shared task, we used only 22 of them (see Section~\ref{sec:data-preproc}).}
covering \nlanguages{} languages
from five language families. Compared to CorefUD 1.2, used last year
\cite{oursharedtask2024}, the release adds three new datasets and two new languages
including Korean, which represents a new language family.
The new datasets are French ANCOR, Hindi HDTB, and Korean ECMT.
In addition, several existing datasets from CorefUD 1.2 were updated.
The data span diverse domains including news, fiction, Bible texts, and Wikipedia articles.
French ANCOR notably introduces transcripts of originally spoken conversational data, which were previously only marginally represented in CorefUD.
Table~\ref{tab:sizes} gives an overview of the datasets and
their sizes. See Appendix~\ref{sec:data-references} for references of the
individual datasets.

One of the goals of the CorefUD project is to encourage research on coreference resolution in languages other than English, particularly those with zero anaphora.
Zero anaphora, or \emph{zero mentions}, occur when a referent (like a subject or object) is implied but not explicitly stated. This is a common feature of pro-drop languages, where verb conjugation often provides enough information to infer the missing pronoun.
In CorefUD, zero mentions are represented as \emph{empty nodes} that are artificially inserted into Universal Dependencies (UD) trees. This allows them to be grouped with other mentions in a coreference chain, just like any other explicitly stated mention.
Although the two newly added languages, Korean and Hindi, are considered pro-drop, the original datasets do not include zero mention annotation.
Therefore, the collection of datasets with zero mentions remains the same as in the previous edition.

Our shared task focuses exclusively on identity coreference. The datasets in the CorefUD collection, however, may include annotations of other relations, such as bridging. Similarly, phenomena like event anaphora and abstract anaphora may be annotated in some datasets but not in others. Because CorefUD is not fully harmonized in terms of annotation guidelines, the precise nature of annotated anaphoric phenomena may vary slightly across corpora. In converting to the CorefUD format, we aim to isolate identity coreference\footnote{We are aware that complete isolation is not possible due to near-identity relations; see \citet{recasensNearIdentity2010}.} while largely preserving the original annotations.

\subsection{New Resources}

\paragraph{French ANCOR} \citep[\francor{};][]{muzerelle-etal-2014-ancor} is a collection of three different corpora of
conversational speech (Accueil\_UBS, OTG, ESLO), annotated for coreference.
Cross-sentence mentions (caused e.g.\ by two speakers speaking simultaneously) are ignored in the conversion from TEI to CorefUD.

\paragraph{Hindi HDTB} \citep[\hihdtb{};][]{mujadia-etal-2016-coreference} is based on the HDTB corpus \citep{palmer2009hindi} annotated with coreference and anaphoric relations and corresponding to the namesake treebank in UD.
However, the coreference corpus does not constitute a strict subset of the UD treebank, as approximately 14\% of its sentences are not included in the UD release. Still, each coreference-annotated document contains at least one sentence that appears in the treebank.
Although the original annotations distinguish \emph{PartOf} relations, these are often merged with identity coreference relations within the same cluster, complicating the separation of identity, bridging, and split-antecedent relations.
As a result, we currently treat all mentions within a cluster as coreferential, without making finer distinctions.
At present, we do not incorporate the manually annotated morpho-syntactic information from the UD treebank; instead, we replace it with automatic parses produced by UDPipe 2.

\paragraph{Korean ECMT} \citep[\koecmt{};][]{nam-etal-2020-effective} is a conversion of the dataset created for the paper
``Effective Crowdsourcing of Multiple Tasks for Comprehensive Knowledge
Extraction'' (ECMT). The original dataset is based on Korean Wikipedia and
KBox with crowdsourced annotations for four information extraction tasks: (1)
entity detection, (2) entity linking, (3) coreference resolution, and (4)
relation extraction. The original dataset seems to contain errors where
distinct entities are incorrectly merged into a single coreference cluster.
The CorefUD conversion did not attempt to fix these errors.

\subsection{Updated Resources}
\label{sec:data-update}

\paragraph{More data} The English GUM corpus (\engum) is now in its version
11, which has approximately 10\% more data. All the other datasets are the
same size as before (except for a few minor changes resulting from annotation
corrections).

\paragraph{New prediction of morphosyntax} For datasets that do not
come with manual morphosyntactic annotation, the UD relations, tags and
features were predicted with newer models for UDPipe (based on UD release
2.15 instead of 2.12). This involves the following ten corpora: Czech PCEDT, English LitBank,
English ParCorFull, German ParCorFull, German PotsdamCC, Hungarian KorKor,
Hungarian SzegedKoref, Lithuanian LCC, Polish PCC, Russian RuCor.

\paragraph{Substantial changes} Re-implementation of conversion from
non-CorefUD formats and/or major revision of the annotation was applied to
Czech PDT (\cspdt) and Hungarian KorKor (\hukorkor). For Czech, the source
dataset is now the PDT part of PDT-C~2.0 (previously it was 1.0), which has
substantial improvements on the surface-syntactic layer. Many other changes
were done in the PDT-to-UD conversion of morphology and syntax; coreference
annotation is unchanged, except for a few corrections. For Hungarian, the
conversion from the native format was almost completely rewritten. Empty
copula nodes are now deleted as required in UD. DROP empty nodes now receive
correct incoming dependency relations (\texttt{nsubj}, \texttt{obj}, or
\texttt{nmod:att}), and there are several other small
improvements.\footnote{More details on the changes can be found in the README
files of the individual corpora.}

\subsection{Data for the Shared Task}
\label{sec:data-preproc}

Compared to the public edition of CorefUD 1.3, the data provided for the shared task participants underwent slight adjustments.\footnote{
Both the shared task data and submissions are available at \url{http://hdl.handle.net/11234/1-5987}.
}

\paragraph{Data reduction}
Firstly, the English and German ParCorFull datasets were excluded from this year's shared task. These datasets are the smallest (their test sets contain less than 900 words, one third of the next smallest test set) and exhibited the largest variance, considerably influencing overall macro-averaged scores.\footnote{Considering eight training runs of the last year's winning system differing in just random initialization, the standard deviation of the ParCorFull development results is more than 10~times larger than the standard deviation of the overall macro-averaged scores and 15~times larger than the standard deviation of the largest dataset.}

Secondly, the development and test sets were reduced to \emph{mini-dev} and \emph{mini-test} sets, respectively.
This change was introduced to lower the computational cost of evaluation while preserving high discriminative power.
Each dev and test set is now capped at 25k words, achieved by randomly sampling complete documents.
The 25k threshold was selected to cut the overall collection size by roughly half, while affecting only a few of the largest corpora and still ensuring reliable and representative results.\footnote{Again considering eight training runs of the last year's winning system differing in just random initialization, capping the large datasets to 25k words increase the standard deviation of the overall macro-averaged percentage results on the development sets by less than +0.03, from 0.296 to 0.324.}

\paragraph{Plaintext format}
For the LLM track, we provide a conversion to a simple plaintext format, along with both the conversion tool and the converted dataset files.

The plaintext format (see Figure~\ref{fig:plaintext-example}) is a plain text file in which each line represents a document, and tokens are separated by spaces.
Coreference annotations are appended to each token after the `\texttt{|}' character.
Each mention, including singletons, is defined by its span boundaries, marked with opening and closing square brackets concatenated with the entity ID.
Empty nodes are prefixed with `\texttt{\#\#}'; if an empty node has a form or lemma in the original data, it is appended immediately after.
Because empty nodes are defined by their syntactic position rather than linear order, each empty node is placed directly after its syntactic parent.
This format does not encode the dependency relation type to the parent, which means it cannot distinguish between multiple empty nodes dependent on the same parent (see Section~\ref{sec:evaluation}).
While this limitation may slightly affect evaluation results, we consider the impact marginal and an acceptable trade-off for preserving the simplicity of the format.

The plaintext format is intentionally less expressive than CoNLL-U and lacks sufficient information for some evaluation metrics (e.g., head match requires mention heads derived from spans using syntactic trees).
To bridge this gap, we provide a backwards conversion tool that restores plaintext annotations to CoNLL-U format, as well as an output cleaner.%
\footnote{The conversion tool and cleaner are available as a single application/Python library on GitHub: \url{https://github.com/ondfa/text2text-coref}}

The cleaner addresses common issues caused by LLM outputs, such as broken annotation structure (e.g., unclosed mentions) or added/removed/modified words.
It first ensures all mentions are properly opened and closed, then uses word-level edit distance to align output documents to the original input.
Empty nodes are ignored in the edit-distance computation, as systems are expected to insert them themselves.
Once the token sequences match exactly, the output annotations can be safely mapped back to the original CoNLL-U files.

\paragraph{Data variants and starting points}
In both tracks, two main variants of the data are provided: gold, and input data.
In addition, participants of the Unconstrained track can choose from three starting points.

\emph{Gold data} includes gold-standard annotations of coreference and empty nodes, intended for fine-tuning and evaluation.
The data are consistent with the CorefUD~1.3 release, retaining manually annotated morpho-syntactic features (for datasets that originally included them), gold empty nodes, and gold coreference annotations.
The only technical modification is the removal of empty nodes' forms in order to align the data with the output of the baseline empty node prediction, which does not predict these forms (see Section~\ref{sec:baseline}).
While the gold train and mini-dev sets were available for download, the gold test set remained secret and were used internally in CodaLab for evaluation.

\emph{Input data} was intended to be processed by participants' systems and
subsequent submission. The following preprocessing was thus performed only on
the mini-dev and mini-test sets. To better simulate a real-world scenario
where no manual linguistic annotation is available, we removed the forms of
empty nodes and replaced the original morpho-syntactic features with the
outputs of UD~2.15 models across all datasets, including those with
originally human-annotated features. Additionally, the gold empty nodes and
coreference annotations were removed, forming the input data for the LLM
track. On the other hand, in line with the setup of the last year's edition,
participants of the Unconstrained track could choose from three different
\emph{starting points} for entering the shared task, with varying degrees of
work required: (1) \emph{Coreference and zeros from scratch} with no
predictions of empty nodes and coreference (practically identical to the
LLM-track variant), (2) \emph{Coreference from scratch} with baseline
predictions of empty nodes, and (3) \emph{Refine the baseline} with baseline
predictions of empty nodes and coreference.

\section{Evaluation Metrics}
\label{sec:evaluation}

The systems participating in the shared task are evaluated using the CorefUD scorer.
In line with previous editions, the primary evaluation score is the CoNLL F$_1$ score, computed with head mention matching and excluding singletons.
To align zero mentions, no longer guaranteed to match one-to-one due to the shift to a more realistic setup introduced last year, we apply a dependency-based matching method.
In addition to the primary metric, we also compute several supplementary scores to support a more comprehensive comparison of the shared task submissions.

\paragraph{Official scorer}
We evaluate participant submissions using the CorefUD scorer%
\footnote{\url{https://github.com/ufal/corefud-scorer}},
specifically the February 2025 version, which remains virtually unchanged from the version used in the previous edition.
The scorer builds on the Universal Anaphora (UA) scorer 2.0 \citep{ua-scorer-2.0},%
\footnote{%
The UA scorer 2.0 merges, reimplements, and extends several earlier tools, including previous versions of the CorefUD scorer.
}
adopting all features relevant to the shared task, including implementations of widely used coreference evaluation metrics.
In contrast to the UA scorer, the CorefUD scorer also supports head matching and a dependency-based method for aligning zero mentions.

The scorer takes two CoNLL-U files as input: the gold file and the predicted file.
Since our plaintext format cannot capture all the information required for evaluation (e.g., mention heads), any LLM output produced in this format must first be restored into CoNLL-U before it can be properly evaluated.

\paragraph{Mention matching}
Due to the limitations of \emph{exact} and \emph{partial} mention matching methods (see \citet{oursharedtask2023} for details), we have settled on the \emph{head match} strategy for the primary evaluation metrics.
In this approach, a gold and predicted mention are considered a match if their heads refer to the same token.%
\footnote{Gold mention heads in the CorefUD data are determined from the dependency tree using the Udapi block \texttt{corefud.MoveHead}.}
Full mention spans are ignored, except in cases where multiple mentions share the same head; in such instances, span information is used to disambiguate them.

However, this approach is not applicable to empty nodes, which frequently occur in zero anaphora.
Predicted counterparts of gold zero mentions may be missing, spurious, or appear at different surface positions within a sentence, even if they serve the same syntactic or semantic role.
To handle this, we devised a \emph{dependency-based method} last year \citep{oursharedtask2024}.
The method aligns predicted and gold zero mentions within the same sentence by maximizing their overlap in enhanced dependency annotations.
It formulates the task as a one-to-one matching in a weighted bipartite graph, where each candidate pair is scored based on how well the predicted zero replicates the gold zero’s dependencies.
Matches that correctly assign both the parent and the dependency type receive higher weights, though the method remains robust even when dependency types are not provided.

\paragraph{Primary score}
As is standard in coreference resolution, we use the CoNLL F$_1$ score \citep{CoNLL-MELA-score,pradhan-etal-2014-scoring} as the primary evaluation metric.
This score is calculated as the unweighted average of the $F_1$ scores from three widely used coreference evaluation measures: MUC \citep{MUC-score}, B$^3$ \citep{Bcubed-score}, and CEAF-e \citep{CEAF-score}.
These metrics offer complementary perspectives: link-based, mention-based, and entity-based, respectively.
As we aim to identify systems with stable performance across all datasets, the final ranking of submissions is determined by the macro-average of CoNLL F$_1$ scores across all mini-test sets in the shared task collection.\footnote{%
The evaluation protocol with macro-averaging CoNLL F$_1$ scores was announced before the start of the development phase and it was used also in previous versions of the shared task.
We think it is the fairest aggregation method.
As alternatives, one could average differences to the baseline or average ranks. The former yields the same final ranking as macro-averaging, while the latter would lead to a single difference: in the LLM track, the winner would be LLM-UWB, despite this system not producing output for one dataset and not covering zero anaphora in some datasets (see Sections~\ref{sec:system-submissions} and~\ref{sec:results}).
}

\paragraph{Supplementary scores}
Beyond the primary CoNLL F$_1$ score, we report its alternative variants based on different mention matching strategies: partial match%
\footnote{Partial match was used as the primary metric in the first edition of the shared task \citep{oursharedtask2022}.}
and exact match.
We also compute the CoNLL score using head match for all mentions, including singletons.

To provide a more comprehensive evaluation, we report the individual coreference metrics comprising the CoNLL score (MUC, B$^3$, and CEAF) as well as other commonly used metrics such as BLANC \citep{BLANC-score} and LEA \citep{LEA-score}.
Furthermore, we include the Mention Overlap Ratio (MOR) to assess mention detection independently of coreference clustering and the anaphor-decomposable score for zero anaphora, both introduced in \citet{oursharedtask2022}.

\section{Participating Systems}
\label{sec:systems}

\subsection{Baseline}
\label{sec:baseline}

As in the previous edition, two baseline systems are provided: one for predicting empty nodes as slots for zero anaphora and another for coreference resolution.
Only participants in the Unconstrained track are permitted to use or build upon these baseline systems.

\paragraph{Empty nodes prediction baseline}

Empty node prediction was introduced as an additional task in last year's
shared task, and it is again part of the shared task this year. To support
participants who wish to focus exclusively on coreference resolution, we
again provide a baseline system for empty nodes prediction. We release
the source code,\footnote{\url{https://github.com/ufal/crac2025_empty_nodes_baseline}}
the trained multilingual model,\footnote{\url{https://www.kaggle.com/models/ufal-mff/crac2025_empty_nodes_baseline/}}
and the mini-dev and mini-test data with predicted empty nodes.

The baseline model architecture is virtually unchanged from last year.
Each input sentence is processed by
a XLM-RoBERTa-large~\citep{conneau-etal-2020-unsupervised}, generating
embeddings for each input word. Then, two candidate empty nodes are predicted
for each word, and passed through three heads: (1)~a binary classification head
predicting whether the candidate is really an empty node or not, (2)~a word-order
prediction head implemented using self-attention selecting the word after which
the empty node should be added, and (3)~a dependency relation prediction head,
which first concatenates the candidate representation and the representation of
the word most probable according to the word-order prediction head, and then
predicts the dependency relation. A single model is trained on a concatenation
of all corpora with empty nodes, sampling every sentence proportionally to the
square root of its corpora size. For a detailed description and a visualization
of the model architecture, see \citet{straka-2024-corpipe}.

\begin{table}[t]
  \centering
  \catcode`! = 13\def!{\itshape}
  \begin{tabular}{lcccc} \toprule
  \textbf{Language} & \kern-.2em\textbf{Recall}\kern-.2em & \kern-.2em\textbf{Precision}\kern-.2em & \kern-.2em\textbf{F1}\kern-.2em & \textbf{!\makecell[c]{2024\\F1}} \\ \midrule
\caancora      & 91.1 & 91.9 & 91.5 & !91.7\\
\cspcedt       & 61.4 & 77.1 & 68.4 & !67.8\\
\cspdt         & 74.9 & 81.0 & 77.8 & !76.2\\
\cuproiel      & 79.0 & 81.0 & 80.0 & !80.2\\
\esancora      & 93.4 & 92.9 & 93.2 & !92.0\\
\grcproiel     & 86.3 & 89.7 & 88.0 & !88.4\\
\hukorkor      & 83.3 & 85.5 & 84.4 & !66.7\\
\huszegedkoref & 87.8 & 88.9 & 88.3 & !90.7\\
\plpcc         & 91.9 & 89.0 & 90.4 & !89.5\\
\tritcc        & 94.0 & 79.8 & 86.3 & !85.8\\

  \bottomrule
  \end{tabular}
  \caption{Empty nodes prediction baseline performance on the minidev sets of CorefUD 1.3 languages containing empty nodes.
    An empty node is considered correct if it has the correct dependency head, dependency relation, and word order.
    For comparison, we also show results from the last year on CorefUD 1.2 dev sets.}
  \label{tab:zeros_baseline}
\end{table}

We intrinsically evaluate the empty node prediction baseline using precision,
recall, and the F1 score, as shown in Table~\ref{tab:zeros_baseline}, where
a prediction is classified as correct only when all of its dependency
head, dependency relation, and word order are correct. For comparison,
we also include the last year's F1 score. This year's results are very
consistent, with the exception of \hukorkor{} showing an increase
of nearly 20 percent points due to improved conversion to the CorefUD
format in CorefUD~1.3 (see Section~\ref{sec:data-update}).

\paragraph{Coreference resolution baseline}
The coreference resolution baseline is the same as in the past three years.
It is based on the multilingual end-to-end neural coreference resolution system by \citet{prazak-etal-2021-multilingual}, which adapts the original end-to-end model of \citet{lee-etal-2017-end}.
The model considers all possible spans up to a predefined maximum length and directly predicts an antecedent for each span.
Since it has no separate mention detection step, it is well suited for datasets that do not annotate singletons.
The baseline uses the mBERT base model as its encoder.

Hereafter, we denote the combination of the two baseline systems as \baseline{} and the coreference resolution baseline applied to gold empty nodes as \baselinegz{}.

\subsection{System Submissions}
\label{sec:system-submissions}
This year, nine systems were submitted to the shared task by six teams:
UWB,\footnote{UWB = University of West Bohemia.}
PUXAI,\footnote{PUXAI refers to the system by Nguyễn Xuân Phúc.}
GLaRef,\footnote{GLaRef = Group Lattice for Reference. Two systems are submitted under this name: GLaRef-CRAC25 and GLaRef-Propp.}
NUST-SEECS,\footnote{NUST-SEECS = National University of Sciences and Technology, School of Electrical Engineering and Computer Science.}
ÚFAL CorPipe,\footnote{ÚFAL CorPipe submitted three variants: CorPipeSingle, CorPipeBestDev, and CorPipeEnsemble.}
and Stanford NLP Group.\footnote{Stanford NLP Group is the creator of the Stanza package.}
For clarity, we distinguish the submissions to the LLM track with the `LLM-' prefix in the following text.

\paragraph{LLM-UWB (hejmanj)}
The UWB team fine-tunes a Llama-3.1-8B model on the official plaintext export of the CoNLL-U
files.  Training is done using QLoRA adaptation. The model is trained to generate the fully tagged document text, including empty nodes, by inserting them directly in the output. For some datasets, they modify the input format to use just a headword for mention representation. Two variants of the model are trained: a simple version using the provided format, but ignoring empty nodes, and an extended version with empty nodes and headword mention representation. Versions for the final submission was selected based on dev set results. The simple version is used for: \cspcedt, \cspdt, \esancora, \grcproiel, \hukorkor, \koecmt, \ltlcc, and \plpcc. For \hboptnk, the model was not properly trained due to very long sequences and inefficient tokenization, and the system failed to meet the output format. Input windows up to 4\,096 tokens are used in training; at inference time, contexts of 2\,048 tokens and outputs of 4\,096 tokens are typical, with occasional extensions to 8\,192/16\,384.  No additional data is used.

\paragraph{LLM-PUXCRAC2025 (PuxAI)}
This system is purely prompt-based, few‐shot coreference resolver combining two closed‐source LLMs (Gemini-Flash-2.0 and Grok-3).  A difficulty‐aware pipeline selects three hardest examples per language, re‐ranks them by two semantic scores, and feeds them plus the test document into the model.  Output chains are post‐processed into CoNLL-U.  No fine‐tuning or extra data is used; the system runs free of charge on public tiers.

\paragraph{LLM-GLaRef-CRAC25 (oseminck)}
The authors fine-tune google/gemma-3-12b-it in two stages: a context‐free end‐to‐end tagger, and a context‐aware variant that processes chunks of sentences (8 or 10 at a time) with preceding context of 500–700 characters.  The best three runs (context‐free, 8sent\_500char, 10sent\_700char) are combined for the final submission.  Training follows QLoRA + prompt tuning + quantization over plaintext inputs; no extra data are used.

\paragraph{LLM-NUST-FewShot (moizsajid)}
This system applies few-shot in-context learning with Gemini 2.5 Pro.  Up to 300k tokens of input are allowed; generation limits are defined by the task.  No fine-tuning or additional data are used.
The system demonstrates that performance scales with the number of examples provided

\paragraph{GLaRef-Propp (antoine.bourgois)}
This work is based on a multi‐stage pipeline built on google/mt5-xl.  Empty nodes are detected first (pro-drop languages only), then mentions with a BiLSTM-CRF, followed by a mention‐pair feedforward coreference scorer.  Windows of up to 512 subwords are used, with sliding overlaps. The three modules contain approximately 54 million trainable parameters and are all fine-tuned solely on CoNLL-U input.

\paragraph{CorPipeSingle (ÚFAL CorPipe)}
The system utilizes a PyTorch re‐implementation of CorPipe24 using google/umt5-xl.  Mentions and links are predicted jointly, but empty nodes are taken from the provided baseline.  The model is trained multi-lingually for 150k gradient updates over 15 epochs; batch sizes of 6–16 sentences with proportional sampling yield the final selected checkpoint.

\paragraph{CorPipeBestDev (ÚFAL CorPipe)}
Same architecture as CorPipeSingle, but instead of one fixed checkpoint, the best checkpoint per treebank (out of 13 models trained with different seeds and sampling) is selected on the mini‐dev sets.

\paragraph{CorPipeEnsemble (ÚFAL CorPipe)}
An ensemble of the top five out of the 13 multilingual umT5-xl models from CorPipeSingle, averaging their predicted mention‐pair probabilities.

\paragraph{Stanza (Stanford NLP Group)}
This work is based on a head-joining efficient word-level conference approach, built on the work of \citet{dobrovolskii-2021-word,doosterlinck-etal-2023-caw,liu-etal-2024-mscaw}. Mentions are first linked by
head words, after which spans are resolved locally through a CNN. Embeddings for mention resolution are initialized via XLM-RoBERTa large, with a sliding
window over the document 512 tokens wide.

\begin{table*}[!t]
\centering
\begin{tabular}{l l l l}
\hline
\textbf{Name} & \textbf{Track} & \textbf{Techniques} \\
\hline
LLM-UWB            & LLM       & FT, LoRA, QLoRA, quant. \\
LLM-PUXCRAC2025    & LLM       & few-shot, re-rank       \\
LLM-GLaRef-CRAC25  & LLM       & FT, prompt-tune, QLoRA, quant. \\
LLM-NUST-FewShot   & LLM       & few-shot in-context     \\
\hline
GLaRef-Propp       & Unconstr. & BiLSTM-CRF + feedforward \\
CorPipeSingle      & Unconstr. & FT multistage           \\
CorPipeBestDev     & Unconstr. & FT + per-treebank select\\
CorPipeEnsemble    & Unconstr. & FT + ensemble           \\
Stanza             & Unconstr. & FT + LoRA               \\
\hline
\end{tabular}
\caption{System names, task tracks, and main techniques.}
\label{tab:sys-tech}
\end{table*}

\begin{table*}[!t]

\centering
\begin{tabular}{l l r r r}
\hline
\textbf{Name} & \textbf{Model} & \textbf{Input ctx. len.} & \textbf{Output tok. len.} & \textbf{\#Params} \\
\hline
LLM-UWB            & Llama-3.1-8B                 & 8,192      & 16,384 & 8 B    \\
LLM-PUXCRAC2025    &  \small\makecell[l]{Gemini-Flash-2.0 \\ Grok-3}  & 1,048,576  & 16,384 & —      \\
LLM-GLaRef-CRAC25  & gemma-3-12b-it               & —          & —      & 12 B   \\
LLM-NUST-FewShot   & Gemini 2.5 Pro               & 300,000    & —      & —      \\
\hline
GLaRef-Propp       & mt5-xl                        & 512        & —      & 54 M   \\
CorPipeSingle      & umT5-xl                       & 512/2,560  & —      & 1.7 B  \\
CorPipeBestDev     & umT5-xl                       & 512/2,560  & —      & 1.7 B  \\
CorPipeEnsemble    & umT5-xl                       & 512/2,560  & —      & 8.6 B  \\
Stanza             & XLM-RoBERTa-L            & 512        & —      & \small\makecell[l]{31M + \\ 560M frozen}  \\
\hline
\end{tabular}
\caption{Models: model name, maximum input context length, maximum new tokens generated, and model sizes.}
\label{tab:context-size}
\end{table*}

\begin{table*}[!t]

\centering
\begin{tabular}{l l r r l}
\hline
\textbf{Name} & \textbf{Empty nodes} & \textbf{Batch size}    & \textbf{Grad ups} & \textbf{Tuned h-params} \\
\hline
LLM-UWB             & \small\makecell[l]{predicted\\ ignored}   & 1                  & ?             & ?                   \\
LLM-PUXCRAC2025     & predicted        & few-shot           & 0           & —                   \\
LLM-GLaRef-CRAC25   & predicted  & ?                & ?             & ?                   \\
LLM-NUST-FewShot      & predicted        & few-shot           & 0             & —                   \\
\hline
GLaRef-Propp    & predicted  & 16,000 mention pairs & ~1.26 M       & batch, epochs       \\
CorPipeSingle   & baseline   & 6 sentences          & 150 k         & sampling mode       \\
CorPipeBestDev  & baseline   & 6 sentences          & 150 k × 13    & same as Single      \\
CorPipeEnsemble & baseline   & 6 sentences          & 150 k × 5     & same as Single      \\
Stanza          & baseline   & 10$\cdot$512-token windows & 367 k          & \small\makecell[l]{learning rate, warmup,\\LoRA params, \dots} \\
\hline
\end{tabular}
\caption{Training configuration: empty-node handling, batch sizes, total gradient updates, and tuned hyperparameters.
GLaRef-Propp used batch size: 16 sentences for empty nodes prediction and mention detection and 16,000 mention pairs for coreference resolution. }
\label{tab:train-config}
\end{table*}

\subsection{System Comparison}\label{sec:system-comparison}
\paragraph{Overview of tables}
Tables~\ref{tab:sys-tech}--\ref{tab:train-config} provide a comprehensive comparison of all nine submissions.  Table~\ref{tab:sys-tech} lists each system’s shared‐task track, primary pretrained backbone, and key methodological components (e.g.\ fine‐tuning, prompt tuning, few‐shot prompting, pipeline modules).  Table~\ref{tab:context-size} details each model’s maximum input context length, maximum new tokens generated at inference, and total number of trainable parameters.  Finally, Table~\ref{tab:train-config} outlines the training regimes: whether models were tuned per language, the batch sizes used, the total number of gradient updates, which hyperparameters were tuned, and how empty nodes were handled.

Although all nine submissions share the same official CoNLL-U training data and target format, they diverge along four main dimensions: modelling paradigm, context capacity, empty node handling, and language‐ or treebank‐specific adaptation.

\paragraph{Modeling paradigms}
There are four contributions in the LLM track and five submissions in the unconstrainted track. The four LLM‐track systems (LLM‐UWB, LLM‐PUXCRAC2025, LLM‐GLaRef-CRAC25, LLM‐NUST-FewShot) treat coreference as a text‐generation or prompt‐answering task. LLM‐UWB and LLM‐GLaRef-CRAC25 perform full fine-tuning (via QLoRA, LoRA, quantization, or prompt tuning) of large open‐source models (Llama-3.1-8B, gemma-3-12b-it), teaching them to output bracketed and empty-node-annotated text. In contrast, LLM‐PUXCRAC2025 and LLM-NUST-FewShot use purely few-shot or in-context prompting on closed-source models (Gemini, Grok), with no parameter updates.

Unconstrained‐track submissions (GLaRef-Propp, CorPipeSingle, CorPipeBestDev, CorPipeEnsemble, Stanza) adopt a more traditional, mention detection -- mention‐pair scoring pipeline. These systems fine-tune XLM-RoBERTa, mT5-xl or umT5-xl in a supervised manner and build clusters via antecedent ranking and transitive closure.

\paragraph{Context capacity and model scale}
The LLM‐track systems exploit the extended context windows of modern LLMs: LLM-UWB up to 8\,192 input / 16\,384 output tokens, LLM-PUXCRAC2025 effectively unlimited (1\,048\,576), and LLM-NUST-FewShot 300\,000 tokens. LLM-GLaRef-CRAC25 similarly benefits from large‐context inference. By contrast, the Unconstrained track systems are limited by standard transformer lengths (512--2\,560 subwords), relying on sliding windows or chunking to cover long documents. Model sizes range from 54 M trainable parameters in GLaRef-Propp’s BiLSTM-CRF modules to 12 B in gemma-3-12b-it, with most systems clustering around 1.7 B–8 B parameters.

\paragraph{Data usage}
All nine systems use only the official CoNLL-U data, with no additional corpora.
Most train a single multilingual model rather than separate per-language models.
The only exception is the CorPipeBestDev system, which picks the best checkpoint per treebank.
In terms of computational cost, only LLM-NUST-FewShot reports a non-zero expense (about \$234.7), while all other systems either report zero cost or rely on university computing resources.

\paragraph{Empty node handling}
Empty nodes are addressed in different ways: (1) predicted end-to-end with a fine-tuned system (LLM-UWB and LLM-GLaRef-CRAC25), (2) predicted end-to-end via in-context learning (LLM-PUXCRAC2025 and LLM-NUST-FewShot), (3) adopted from the shared task's baseline (CorPipe variants, Stanza), or (4) predicted with a custom model (GLaRef-Propp).
The LLM-based systems relied on the serialized format, which represents empty nodes using `\texttt{\#\#}' markers (see Figure~\ref{fig:plaintext-example}).
These varied approaches reflect different assumptions about the importance and difficulty of modeling zero-anaphora phenomena.

\paragraph{Language/treebank specialization and ensembling}
Most systems train a single multilingual model for all languages (LLM-UWB, LLM-PUXCRAC2025, GLaRef-CRAC25, NUST-FewShot, GLaRef-Propp, CorPipeSingle, Stanza).
Only CorPipeBestDev and CorPipeEnsemble select or combine checkpoints: CorPipeBestDev picks the best of 195 (13 models $\cdot$ 15 epochs) multilingual checkpoints for each corpus, while Ensemble averages the top five multilingual models.
Neither LLM-UWB nor LLM-GLaRef-CRAC25 employ per-language tuning, favoring a unified model.
The few-shot systems dynamically adapt to each input via prompt construction but do not explicitly retrain per language.

In sum, the task saw a spectrum from lightweight, prompt‐only solutions on closed LLM APIs to heavyweight, quantized fine-tuned open models, and from end-to-end generation of annotations to modular neural-pipeline architectures.

\section{Results and Comparison}
\label{sec:results}

\paragraph{Main results}
The main results are summarized in Table~\ref{tab:main-results}.
LLM-GLaRef-CRAC25 and CorPipeEnsemble are the top-performing systems in the LLM and Unconstrained tracks, respectively, outperforming all other submissions in their respective tracks according to the primary metric. Both systems also achieve the best results within their track when evaluated with alternative mention matching strategies: partial match, exact match, and head match including singletons.

The LLM track exhibits tighter competition, with performance differences between systems significantly smaller than in the Unconstrained track. Excluding the baseline system, the standard deviation of the head match score in the Unconstrained track is 5.53, compared to just 1.27 in the LLM track.
This higher level of competition is also reflected in the progression of scores over time, as shown in Figure~\ref{fig:codalab-evol} in Appendix~\ref{sec:codalab-evol}, which tracks the evolution of primary scores for individual submissions during the evaluation phase of the shared task.

Comparing across the tracks, all LLMs could beat the non-LLM baseline system.
However, we have to admit that %
in this shared task the best LLM solution
fell behind the best non-LLM system by a large margin of almost 13 points.
For simplicity, we will be comparing the submissions from both tracks jointly in the remainder of this section.

\paragraph{Secondary metrics}
The secondary metrics in Table~\ref{tab:secondary-metrics} reveal a similar trend as the primary metric: the ÚFAL CorPipe system consistently outperforms all other submissions.
The most striking pattern is the pronounced contrast between the CorPipe systems and the remaining entries, particularly the LLM-based ones, in terms of the precision–recall balance across individual coreference metrics.
While CorPipe systems maintain relatively small gaps between precision and recall, the other systems consistently show much higher precision than recall.
This indicates that CorPipe systems are substantially more effective at capturing and following the coreference annotation guidelines reflected in the data.

\paragraph{Comparison across datasets}
Both Table~\ref{tab:all-langs} and Figure~\ref{fig:alllangs} present CoNLL F$_1$ scores of all systems across the datasets.
To make patterns more visible, the datasets in Figure~\ref{fig:alllangs} are ordered from left to right by the decreasing performance of the top system, CorPipeEnsemble.
For roughly the lower-performing half of the datasets, the performance gap between CorPipe and the other systems tends to be larger, and their scores are more varied, suggesting that these datasets pose greater challenges for coreference resolution.

Interestingly, CorPipeEnsemble was outperformed on two datasets: \enlitbank{} by LLM-UWB, and \hboptnk{} by LLM-NUST-FewShot.
The latter is particularly striking: on Ancient Hebrew, LLM-NUST-FewShot surpassed CorPipeEnsemble by 10 points, despite ranking among the weakest systems on many other datasets.
While the exact cause of this anomaly remains unclear, a closer analysis shows that LLM-NUST-FewShot produced almost exactly the same number of non-singleton mentions as in the gold data (2,327 vs. 2,312), whereas %
all other system produced less mentions.

The zero score of LLM-UWB on \hboptnk{} is in line with their fine-tuning failure described in Section~\ref{sec:system-submissions}.

\paragraph{Performance on zero mentions}
Table~\ref{tab:all-langs-zero} shows system performance on datasets containing zero mentions, evaluated using the anaphor-decomposable score for zero anaphora.
Two observations stand out.

First, LLM-UWB fails to predict any zero mentions for all but two of these datasets.
This is likely because several of these datasets substantially overlap with those for which the authors used an LLM variant fine-tuned on data where empty nodes had been excluded.

Second, on \hukorkor{}, both the winning system and the baseline outperform their counterparts from last year's edition by 8 and 10 percentage points, respectively.
The winning system's score is now closer to its performance on the other Hungarian dataset, \huszegedkoref{}.
These gains are consistent with the improved intrinsic performance of the empty-node prediction baseline for this dataset (see Section~\ref{sec:baseline}), resulting from fixes to its conversion pipeline described in Section~\ref{sec:data-update}.

\paragraph{Comparison over years}
Having organized this shared task for the fourth consecutive year, it is particularly interesting to examine how it has contributed to advancing the state of the art in multilingual coreference resolution.
While the datasets and certain aspects of the task have evolved each year, one constant has been the coreference baseline system, which is simply retrained annually on the updated data.
This stability allows us to track progress by comparing the best-performing system each year against the baseline.

The relative improvement over the baseline showed a promising upward trend in previous editions: +21\% in 2022, +31\% in 2023, and +39\% in 2024 \citep{oursharedtask2024}.
This year, however, the improvement stands at +35\%, marking a slight break in the upward trajectory.
This drop is caused by the exclusion of two very small datasets from the test set,
 where the improvement over baseline has been exceptionally high last year
 (+47\% in de\_parcorfull and +108\% in en\_parcorful) perhaps by chance.
Still, the results show that systems continue to deliver strong performance even as the task grows more diverse and challenging.

\begin{table*}\centering
\begin{tabular}{@{}l r r@{~~}l r@{~~}l r@{~~}l @{}}\toprule
                  & \MC{5}{excluding singletons} & \MC{2}{with singletons}\\\cmidrule(lr){2-6}\cmidrule(l){7-8}
\bf system        & \bf head-match & \MC{2}{\bf partial-match} & \MC{2}{\bf exact-match} & \MC{2}{\bf head-match}\\\midrule
LLM-GLaRef-CRAC25 & \bf 62.96 & \bf 61.66 & (-1.30) & \bf 58.98 & (-3.98) & \bf 65.61 & (+2.66)\\
LLM-NUST-FewShot  &     61.74 &     61.14 & (-0.60) &     56.34 & (-5.40) &     63.44 & (+1.69)\\
LLM-PUXCRAC2025   &     60.09 &     59.68 & (-0.41) &     55.22 & (-4.87) &     54.77 & (-5.32)\\
LLM-UWB           &     59.84 &     59.55 & (-0.29) &     38.81 &(-21.03) &     62.77 & (+2.93)\\\midrule
CorPipeEnsemble   & \bf 75.84 & \bf 74.90 & (-0.94) & \bf 72.76 & (-3.08) & \bf 78.33 & (+2.49)\\
CorPipeBestDev    &     75.06 &     74.08 & (-0.98) &     71.97 & (-3.10) &     77.63 & (+2.57)\\
CorPipeSingle     &     74.75 &     73.74 & (-1.01) &     71.53 & (-3.23) &     77.43 & (+2.68)\\
Stanza            &     67.81 &     67.03 & (-0.78) &     64.68 & (-3.13) &     70.64 & (+2.83)\\
GLaRef-Propp      &     61.57 &     60.72 & (-0.85) &     58.43 & (-3.14) &     65.28 & (+3.70)\\
\baselinegz       &     58.18 &     57.75 & (-0.42) &     56.48 & (-1.69) &     49.88 & (-8.29)\\
\baseline         &     56.01 &     55.58 & (-0.43) &     54.24 & (-1.77) &     47.88 & (-8.13)\\
\midrule
\sys{Winner-2023}   &    74.90 &      73.33 & (-1.57) &     71.46 & (-3.44) &     76.82 & (+1.91)\\
\sys{Winner-2024}   &    73.90 &      72.19 & (-1.71) &     69.86 & (-4.04) &     75.65 & (+1.75)\\
\sys{Baseline-2023} &    56.96 &      56.28 & (-0.68) &     54.75 & (-2.21) &     49.32 & (-7.64)\\
\sys{Baseline-2024} &    53.16 &      52.48 & (-0.68) &     51.26 & (-1.90) &     46.45 & (-6.71)\\
\bottomrule\end{tabular}

\caption{Main results: the CoNLL F$_1$ score macro-averaged over all datasets.
The table shows the primary metric (head-match excluding singletons)
 and three alternative metrics:
 partial-match excluding singletons,
 exact-match excluding singletons and
 head-match with singletons.
A difference relative to the primary metric is reported in parenthesis.
The top section shows the LLM track, below is the Unconstrained track.
The best score in each column and each of these two sections is in bold.
The systems are ordered by the primary metric.
The last four rows showing the winner and baseline results from CRAC~2023 and 2024
 are copied from the last year Findings \citep{oursharedtask2024},
 and thus are not directly comparable with the rest of the table
 because both the test and training data have been changed (CorefUD~1.1 vs. 1.2 vs. 1.3).
Similar notes apply to the following tables.
}
\label{tab:main-results}
\end{table*}

\begin{table*}\centering
\begin{tabular}{@{}l cccccc @{}}\toprule
\bf system        & \bf MUC &\bf B$^3$ &\bf CEAF-e &\bf BLANC &\bf LEA &\bf MOR\\\midrule
CorPipeEnsemble   &  {\bf 81} / {\bf 82} / {\bf 82}  &  {\bf 73} / {\bf 75} / {\bf 74}  &  {\bf 74} / {\bf 70} / {\bf 72}  &       72  / {\bf 75} / {\bf 73}  &  {\bf 70} / {\bf 73} / {\bf 71}  &       81  /      82  / {\bf 81} \\
CorPipeBestDev    &       81  /      81  /      81   &       72  /      74  /      73   &       73  /      70  /      71   &  {\bf 72} /      74  /      73   &       70  /      71  /      70   &  {\bf 81} /      81  /      81  \\
CorPipeSingle     &       81  /      81  /      81   &       72  /      73  /      72   &       72  /      70  /      71   &       72  /      73  /      72   &       69  /      71  /      70   &       80  /      81  /      80  \\
Stanza            &       72  /      80  /      76   &       62  /      70  /      65   &       62  /      64  /      63   &       61  /      70  /      64   &       59  /      67  /      62   &       70  /      83  /      75  \\
LLM-GLaRef-CRAC25 &       67  /      76  /      71   &       55  /      67  /      60   &       55  /      61  /      58   &       54  /      67  /      59   &       51  /      64  /      56   &       64  /      79  /      71  \\
LLM-NUST-FewShot  &       66  /      73  /      69   &       58  /      65  /      60   &       52  /      65  /      56   &       57  /      65  /      58   &       56  /      62  /      57   &       59  /      79  /      66  \\
GLaRef-Propp      &       69  /      76  /      72   &       56  /      62  /      58   &       49  /      62  /      55   &       56  /      62  /      57   &       52  /      58  /      55   &       57  /      78  /      65  \\
LLM-PUXCRAC2025   &       64  /      72  /      68   &       54  /      63  /      57   &       52  /      61  /      55   &       53  /      62  /      56   &       51  /      59  /      54   &       56  /      80  /      65  \\
LLM-UWB           &       60  /      74  /      65   &       53  /      67  /      57   &       53  /      64  /      57   &       48  /      67  /      53   &       50  /      64  /      55   &       42  /      81  /      53  \\
\baselinegz       &       61  /      76  /      68   &       48  /      63  /      54   &       49  /      58  /      52   &       48  /      64  /      54   &       45  /      59  /      50   &       55  / {\bf 87} /      66  \\
\baseline         &       58  /      75  /      65   &       45  /      62  /      52   &       47  /      57  /      51   &       44  /      63  /      50   &       42  /      58  /      48   &       53  /      86  /      65  \\
\bottomrule\end{tabular}
\caption{Recall / Precision / F1 for individual secondary metrics.
All scores macro-averaged over all datasets.
}
\label{tab:secondary-metrics}
\end{table*}

\newcommand*\rot{\rotatebox{90}}
\begin{table*}\centering
\resizebox{\textwidth}{!}{
\begin{tabular}{@{}l r@{~~~}r@{~~~}r@{~~~}r@{~~~}r@{~~~}r@{~~~}r@{~~~}r@{~~~}r@{~~~}r@{~~~}r@{~~~}r@{~~~}r@{~~~}r@{~~~}r@{~~~}r@{~~~}r@{~~~}r@{~~~}r@{~~~}r@{~~~}r@{~~~}r@{}}\toprule
\bf system        & \rot\caancora & \rot\cspcedt & \rot\cspdt & \rot\cuproiel & \rot\depotsdamcc & \rot\engum & \rot\enlitbank & \rot\esancora & \rot\francor & \rot\frdemocrat & \rot\grcproiel & \rot\hboptnk & \rot\hihdtb & \rot\hukorkor & \rot\huszegedkoref & \rot\koecmt & \rot\ltlcc & \rot\nobokmaalnarc & \rot\nonynorsknarc & \rot\plpcc & \rot\rurucor & \rot\tritcc\\\midrule
CorPipeEnsemble   &\bf  82.9 &\bf  77.1 &\bf  80.7 &\bf  65.5 &\bf  73.0 &\bf  76.1 &     81.8 &\bf  84.5 &\bf  76.3 &\bf  71.8 &\bf  74.5 &     69.8 &\bf  77.7 &\bf  68.6 &\bf  71.0 &\bf  69.9 &\bf  77.2 &\bf  78.2 &\bf  76.3 &\bf  80.2 &     84.2 &     71.2\\
CorPipeBestDev    &     82.0 &     76.3 &     80.4 &     62.8 &     72.6 &     75.9 &     81.3 &     83.8 &     75.9 &     69.9 &     74.3 &     68.3 &     77.5 &     68.3 &     70.5 &     69.3 &     76.0 &     77.1 &     74.0 &     79.9 &\bf  84.8 &     70.4\\
CorPipeSingle     &     82.5 &     76.2 &     80.1 &     63.0 &     72.8 &     75.2 &     80.8 &     84.1 &     75.8 &     70.3 &     74.4 &     66.1 &     76.5 &     67.3 &     69.7 &     68.9 &     75.8 &     76.2 &     73.6 &     79.4 &     84.2 &\bf  71.6\\
Stanza            &     79.5 &     72.7 &     75.1 &     40.8 &     67.3 &     69.0 &     74.8 &     80.4 &     67.5 &     62.5 &     54.9 &     62.1 &     74.2 &     60.0 &     64.6 &     67.7 &     72.8 &     72.4 &     71.7 &     73.0 &     80.8 &     47.8\\
LLM-GLaRef-CRAC25 &     73.5 &     65.1 &     71.3 &     58.2 &     59.6 &     58.7 &     69.0 &     74.4 &     66.7 &     60.4 &     65.8 &     44.0 &     56.4 &     52.5 &     59.8 &     63.0 &     62.5 &     64.7 &     61.6 &     72.5 &     68.8 &     56.2\\
LLM-NUST-FewShot  &     60.9 &     51.4 &     54.3 &     58.5 &     48.7 &     69.8 &     70.4 &     61.8 &     71.9 &     57.6 &     57.9 &\bf  80.2 &     71.3 &     43.5 &     52.3 &     66.0 &     59.2 &     72.8 &     68.9 &     70.8 &     71.4 &     39.0\\
GLaRef-Propp      &     68.1 &     61.7 &     66.6 &     39.1 &     61.2 &     61.9 &     70.0 &     69.1 &     65.1 &     66.1 &     51.3 &     58.8 &     69.5 &     50.9 &     60.1 &     60.6 &     57.6 &     67.1 &     66.3 &     68.0 &     71.5 &     44.3\\
LLM-PUXCRAC2025   &     68.0 &     56.9 &     63.0 &     43.7 &     57.4 &     61.7 &     69.1 &     70.5 &     63.8 &     61.5 &     47.9 &     45.3 &     66.8 &     50.6 &     61.6 &     50.3 &     65.3 &     65.2 &     63.0 &     66.5 &     67.6 &     56.1\\
LLM-UWB           &     79.2 &     61.0 &     68.2 &     25.3 &     67.6 &     73.6 &\bf  84.0 &     73.6 &     58.6 &     49.1 &     47.6 &      0.0 &     75.8 &     38.9 &     67.3 &     68.3 &     63.4 &     73.8 &     72.0 &     64.5 &     80.1 &     24.3\\
\baselinegz       &     68.8 &     69.5 &     67.9 &     29.5 &     55.7 &     61.6 &     66.0 &     71.0 &     63.8 &     55.0 &     29.4 &     31.0 &     66.8 &     47.1 &     54.3 &     64.3 &     65.3 &     62.5 &     63.0 &     68.1 &     67.6 &     51.7\\
\baseline         &     68.0 &     56.9 &     63.0 &     26.3 &     55.7 &     61.7 &     66.0 &     70.5 &     63.8 &     55.0 &     28.5 &     31.0 &     66.8 &     43.2 &     54.5 &     50.3 &     65.3 &     62.5 &     63.0 &     66.5 &     67.6 &     45.9\\
\bottomrule\end{tabular}
}
\caption{Results for individual languages in the primary metric (CoNLL F$_1$).
}
\label{tab:all-langs}
\end{table*}

\begin{figure*}\centering
\begin{tikzpicture}
  \begin{axis}[
    ylabel={CoNLL F$_1$},
    ymin=20, ymax=85,
    width=\textwidth, height=19cm,
    xtick={1, 2, 3, 4, 5, 6, 7, 8, 9, 10, 11, 12, 13, 14, 15, 16, 17, 18, 19, 20, 21, 22},
    xticklabels={\rot\esancora, \rot\rurucor, \rot\caancora, \rot\enlitbank, \rot\cspdt, \rot\plpcc, \rot\nobokmaalnarc, \rot\hihdtb, \rot\ltlcc, \rot\cspcedt, \rot\francor, \rot\nonynorsknarc, \rot\engum, \rot\grcproiel, \rot\depotsdamcc, \rot\frdemocrat, \rot\tritcc, \rot\huszegedkoref, \rot\koecmt, \rot\hboptnk, \rot\hukorkor, \rot\cuproiel},
    legend style={at={(0.5, -0.2)}, anchor=north},
    legend columns=3,
    ymajorgrids=true,
    grid style=dashed
  ]
    \addplot [solid, color=blue, mark=*] coordinates {
      (1, 84.5)
      (2, 84.2)
      (3, 82.9)
      (4, 81.8)
      (5, 80.7)
      (6, 80.2)
      (7, 78.2)
      (8, 77.7)
      (9, 77.2)
      (10, 77.1)
      (11, 76.3)
      (12, 76.3)
      (13, 76.1)
      (14, 74.5)
      (15, 73.0)
      (16, 71.8)
      (17, 71.2)
      (18, 71.0)
      (19, 69.9)
      (20, 69.8)
      (21, 68.6)
      (22, 65.5)
    };
    \addlegendentry{CorPipeEnsemble}
    \addplot [solid, color=blue!60!black, mark=o] coordinates {
      (1, 83.8)
      (2, 84.8)
      (3, 82.0)
      (4, 81.3)
      (5, 80.4)
      (6, 79.9)
      (7, 77.1)
      (8, 77.5)
      (9, 76.0)
      (10, 76.3)
      (11, 75.9)
      (12, 74.0)
      (13, 75.9)
      (14, 74.3)
      (15, 72.6)
      (16, 69.9)
      (17, 70.4)
      (18, 70.5)
      (19, 69.3)
      (20, 68.3)
      (21, 68.3)
      (22, 62.8)
    };
    \addlegendentry{CorPipeBestDev}
    \addplot [solid, color=blue!60!green, mark=square*] coordinates {
      (1, 84.1)
      (2, 84.2)
      (3, 82.5)
      (4, 80.8)
      (5, 80.1)
      (6, 79.4)
      (7, 76.2)
      (8, 76.5)
      (9, 75.8)
      (10, 76.2)
      (11, 75.8)
      (12, 73.6)
      (13, 75.2)
      (14, 74.4)
      (15, 72.8)
      (16, 70.3)
      (17, 71.6)
      (18, 69.7)
      (19, 68.9)
      (20, 66.1)
      (21, 67.3)
      (22, 63.0)
    };
    \addlegendentry{CorPipeSingle}
    \addplot [solid, color=red, mark=triangle*] coordinates {
      (1, 80.4)
      (2, 80.8)
      (3, 79.5)
      (4, 74.8)
      (5, 75.1)
      (6, 73.0)
      (7, 72.4)
      (8, 74.2)
      (9, 72.8)
      (10, 72.7)
      (11, 67.5)
      (12, 71.7)
      (13, 69.0)
      (14, 54.9)
      (15, 67.3)
      (16, 62.5)
      (17, 47.8)
      (18, 64.6)
      (19, 67.7)
      (20, 62.1)
      (21, 60.0)
      (22, 40.8)
    };
    \addlegendentry{Stanza}
    \addplot [dashed, color=green!60!black, mark=diamond*] coordinates {
      (1, 74.4)
      (2, 68.8)
      (3, 73.5)
      (4, 69.0)
      (5, 71.3)
      (6, 72.5)
      (7, 64.7)
      (8, 56.4)
      (9, 62.5)
      (10, 65.1)
      (11, 66.7)
      (12, 61.6)
      (13, 58.7)
      (14, 65.8)
      (15, 59.6)
      (16, 60.4)
      (17, 56.2)
      (18, 59.8)
      (19, 63.0)
      (20, 44.0)
      (21, 52.5)
      (22, 58.2)
    };
    \addlegendentry{LLM-GLaRef-CRAC25}
    \addplot [dashed, color=orange, mark=pentagon*] coordinates {
      (1, 61.8)
      (2, 71.4)
      (3, 60.9)
      (4, 70.4)
      (5, 54.3)
      (6, 70.8)
      (7, 72.8)
      (8, 71.3)
      (9, 59.2)
      (10, 51.4)
      (11, 71.9)
      (12, 68.9)
      (13, 69.8)
      (14, 57.9)
      (15, 48.7)
      (16, 57.6)
      (17, 39.0)
      (18, 52.3)
      (19, 66.0)
      (20, 80.2)
      (21, 43.5)
      (22, 58.5)
    };
    \addlegendentry{LLM-NUST-FewShot}
    \addplot [solid, color=green!60!black, mark=*] coordinates {
      (1, 69.1)
      (2, 71.5)
      (3, 68.1)
      (4, 70.0)
      (5, 66.6)
      (6, 68.0)
      (7, 67.1)
      (8, 69.5)
      (9, 57.6)
      (10, 61.7)
      (11, 65.1)
      (12, 66.3)
      (13, 61.9)
      (14, 51.3)
      (15, 61.2)
      (16, 66.1)
      (17, 44.3)
      (18, 60.1)
      (19, 60.6)
      (20, 58.8)
      (21, 50.9)
      (22, 39.1)
    };
    \addlegendentry{GLaRef-Propp}
    \addplot [dashed, color=purple, mark=o] coordinates {
      (1, 70.5)
      (2, 67.6)
      (3, 68.0)
      (4, 69.1)
      (5, 63.0)
      (6, 66.5)
      (7, 65.2)
      (8, 66.8)
      (9, 65.3)
      (10, 56.9)
      (11, 63.8)
      (12, 63.0)
      (13, 61.7)
      (14, 47.9)
      (15, 57.4)
      (16, 61.5)
      (17, 56.1)
      (18, 61.6)
      (19, 50.3)
      (20, 45.3)
      (21, 50.6)
      (22, 43.7)
    };
    \addlegendentry{LLM-PUXCRAC2025}
    \addplot [dashed, color=teal, mark=square*] coordinates {
      (1, 73.6)
      (2, 80.1)
      (3, 79.2)
      (4, 84.0)
      (5, 68.2)
      (6, 64.5)
      (7, 73.8)
      (8, 75.8)
      (9, 63.4)
      (10, 61.0)
      (11, 58.6)
      (12, 72.0)
      (13, 73.6)
      (14, 47.6)
      (15, 67.6)
      (16, 49.1)
      (17, 24.3)
      (18, 67.3)
      (19, 68.3)
      (20, 0.0)
      (21, 38.9)
      (22, 25.3)
    };
    \addlegendentry{LLM-UWB}
    \addplot [solid, color=gray, mark=triangle*] coordinates {
      (1, 71.0)
      (2, 67.6)
      (3, 68.8)
      (4, 66.0)
      (5, 67.9)
      (6, 68.1)
      (7, 62.5)
      (8, 66.8)
      (9, 65.3)
      (10, 69.5)
      (11, 63.8)
      (12, 63.0)
      (13, 61.6)
      (14, 29.4)
      (15, 55.7)
      (16, 55.0)
      (17, 51.7)
      (18, 54.3)
      (19, 64.3)
      (20, 31.0)
      (21, 47.1)
      (22, 29.5)
    };
    \addlegendentry{\baselinegz}
    \addplot [solid, color=black, mark=diamond*] coordinates {
      (1, 70.5)
      (2, 67.6)
      (3, 68.0)
      (4, 66.0)
      (5, 63.0)
      (6, 66.5)
      (7, 62.5)
      (8, 66.8)
      (9, 65.3)
      (10, 56.9)
      (11, 63.8)
      (12, 63.0)
      (13, 61.7)
      (14, 28.5)
      (15, 55.7)
      (16, 55.0)
      (17, 45.9)
      (18, 54.5)
      (19, 50.3)
      (20, 31.0)
      (21, 43.2)
      (22, 26.3)
    };
    \addlegendentry{\baseline}
  \end{axis}
\end{tikzpicture}
\caption{Plot with results for individual languages in the primary metric (CoNLL F$_1$).
This plot shows the same information as Table~\ref{tab:all-langs},
but languages are sorted according to the performance of the best system
and LLM-based systems are shown with dashed lines.}
\label{fig:alllangs}
\end{figure*}
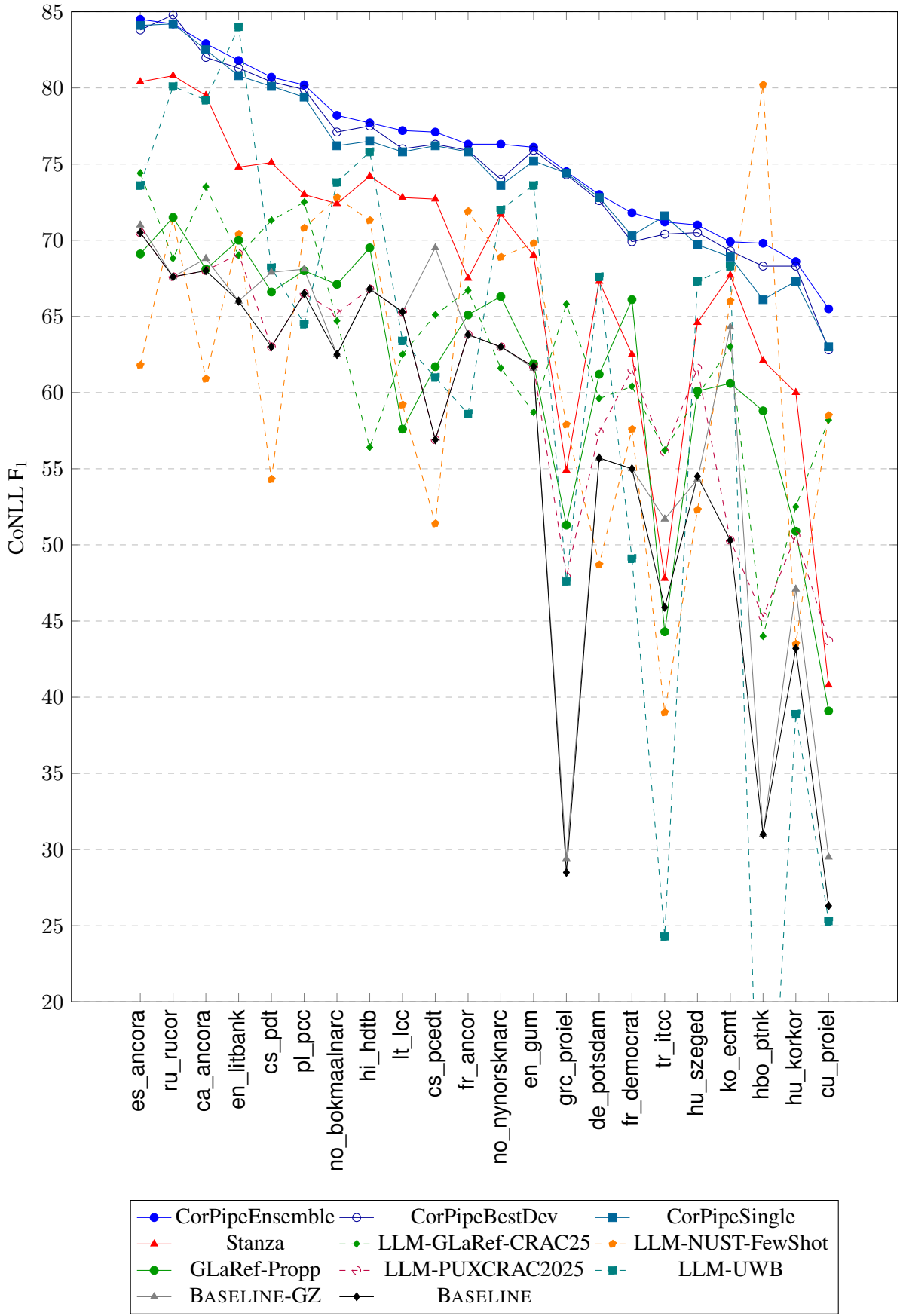

\begin{table*}\centering
\resizebox{\textwidth}{!}{
\begin{tabular}{@{}l r@{~~~}r@{~~~}r@{~~~}r@{~~~}r@{~~~}r@{~~~}r@{~~~}r@{~~~}r@{~~~}r@{}}\toprule
\bf system        & \rot\caancora & \rot\cspdt & \rot\cspcedt & \rot\cuproiel & \rot\esancora & \rot\grcproiel & \rot\hukorkor & \rot\huszegedkoref & \rot\plpcc & \rot\tritcc\\\midrule
CorPipeEnsemble   &  {\bf 91} / {\bf 87} / {\bf 89}  &       82  /      86  / {\bf 84}  &       61  /      79  /      69   &  {\bf 77} /      80  / {\bf 79}  &       93  / {\bf 92} / {\bf 92}  &       87  /      87  /      87   &       65  /      81  /      72   &  {\bf 85} / {\bf 73} / {\bf 78}  &       93  /      84  / {\bf 89}  &  {\bf 84} /      83  / {\bf 84} \\
CorPipeBestDev    &       90  /      87  /      88   &       82  /      85  /      84   &       60  /      77  /      68   &       76  /      79  /      78   &  {\bf 93} /      91  /      92   &  {\bf 87} /      88  / {\bf 88}  &       66  /      82  /      73   &       83  /      70  /      76   &       93  /      84  /      88   &       84  /      82  /      83  \\
CorPipeSingle     &       90  /      86  /      88   &       81  /      85  /      83   &       61  /      78  /      68   &       77  /      79  /      78   &       93  /      92  /      92   &  {\bf 87} /      88  /      88   &       63  / {\bf 83} /      72   &       83  /      70  /      76   &  {\bf 94} /      83  /      88   &       84  /      82  /      83  \\
Stanza            &       87  /      86  /      86   &       77  / {\bf 88} /      82   &       52  /      84  /      65   &       63  /      69  /      66   &       91  /      91  /      91   &       80  /      84  /      82   &       59  /      83  /      69   &       74  /      70  /      72   &       91  /      81  /      86   &       57  /      83  /      67  \\
LLM-GLaRef-CRAC25 &       81  /      84  /      82   &       75  /      81  /      78   &       56  /      67  /      61   &       77  /      79  /      78   &       83  /      89  /      86   &       85  /      87  /      86   &       52  /      68  /      59   &       66  /      65  /      65   &       84  /      83  /      84   &       75  /      75  /      75  \\
LLM-NUST-FewShot  &       53  /      82  /      64   &       55  /      79  /      65   &       35  /      81  /      48   &       74  / {\bf 82} /      78   &       56  /      91  /      69   &       59  / {\bf 89} /      71   &       23  /      83  /      36   &       25  /      63  /      36   &       72  /      86  /      79   &       29  /      63  /      40  \\
GLaRef-Propp      &       80  /      80  /      80   &       74  /      83  /      78   &       48  /      63  /      54   &       49  /      56  /      53   &       84  /      87  /      86   &       70  /      74  /      72   &       51  /      70  /      59   &       66  /      66  /      66   &       84  /      82  /      83   &       60  /      83  /      70  \\
LLM-PUXCRAC2025   &       79  /      75  /      77   &       34  /      82  /      48   &        9  / {\bf 93} /      17   &       39  /      53  /      45   &       88  /      87  /      87   &       82  /      60  /      69   &       50  /      48  /      49   &       73  /      49  /      59   &       86  /      78  /      82   &       50  / {\bf 93} /      65  \\
LLM-UWB           &       83  /      82  /      82   &        0  /       0  /       0   &        0  /       0  /       0   &        0  /       0  /       0   &        0  /       0  /       0   &        0  /       0  /       0   &        0  /       0  /       0   &       71  /      73  /      72   &        0  /       0  /       0   &        0  /       0  /       0  \\
\baselinegz       &       84  /      83  /      84   &  {\bf 83} /      85  /      84   &  {\bf 76} /      81  / {\bf 79}  &       61  /      71  /      66   &       89  /      90  /      90   &       64  /      67  /      66   &  {\bf 73} /      76  / {\bf 74}  &       54  /      59  /      56   &       89  / {\bf 87} /      88   &       79  /      81  /      80  \\
\baseline         &       79  /      75  /      77   &       34  /      82  /      48   &        9  / {\bf 93} /      17   &       52  /      62  /      57   &       88  /      87  /      87   &       62  /      67  /      64   &       56  /      63  /      59   &       54  /      57  /      55   &       86  /      78  /      82   &       71  /      73  /      72  \\
\midrule
\sys{Winner-2023}  &  93 / 92 / 92  &  91 / 92 / 92  &  87 / 88 / 87 & --  &  94 /      95  / 95  & -- &  82 /      89  / 85  &  88 /      70  /      78   &       75  /      69  /      72 & --  \\
\sys{Winner-2024} &       88  /      85  /      86   &       77  /      82  /      80   &       59  /      74  /      66   &   75 / 78 / 76  &  90 / 92 / 91  &  84 / 88 / 86  &       56  /      75  /      64   &  83 /      68  /      75   &  90 /      84  /      87   &  83 /      80  /      82  \\
\sys{Baseline-2023}&       82  /      82  /      82   &       81  /      84  /      82   &       77  /      81  /      79   & -- &       87  /      88  /      87   &  -- &      60  /      68  /      64   &       61  /      57  /      59   &       50  / 80 /      62  & -- \\
\sys{Baseline-2024}&       79  /      76  /      77   &       70  /      74  /      72   &       55  /      69  /      61   &       52  /      62  /      56   &       83  /      83  /      83   &       63  /      70  /      66   &       41  /      61  /      49   &       49  /      57  /      53   &       85  /      78  /      82   &       68  /      71  /      70  \\
\bottomrule\end{tabular}
}
\caption{Recall / Precision / F1 for anaphor-decomposable score of coreference resolution on zero anaphors across individual languages.
Only datasets containing anaphoric zeros are listed (\engum{} excluded as all zeros in its test set are non-anaphoric).
Note that these scores are directly comparable to neither the CoNLL score nor the supplementary scores calculated with respect to whole entities in Table~\ref{tab:secondary-metrics}.
}
\label{tab:all-langs-zero}
\end{table*}

\paragraph{Further analysis}
Similarly to previous years, we provide several additional tables in the
appendices to shed more light on the differences between the submitted
systems.

Tables~\ref{tab:upos-entity}--\ref{tab:upos-mention} show results factorized
according to the different universal part of speech tags (UPOS) in the
mention heads. 

Tables \ref{tab:stats-entities}--\ref{tab:stats-details} show various
statistics on the entities and mentions in a concatenation of all the test
sets. Note that such statistics are mostly influenced by larger datasets.

\paragraph{Differences between LLM and Unconstrained}

The main novelty in this year's shared task setup was the support for LLM approaches to coreference resolution. 
As mentioned in the Main Results above, the performance of the LLM participating systems
 is worse than the best Unconstrained system (CorPipe) by a large margin
 (with only two datasets where an LLM system outperforms all Unconstrained systems).
In addition, some LLM systems seem to be sensitive to particular datasets:
 there are
 dramatic drops in performance (see e.g.\ the performance declines for \texttt{grc\_proiel}, \texttt{tr\_itcc}, \texttt{hbo\_ptnk}, and \texttt{cu\_proiel} in Figure~\ref{fig:alllangs}).

However, it would be premature to conclude that LLMs are not a promising solution for coreference resolution. First, this would contradict everyday experience with public LLMs, which seem to handle coreference-related phenomena relatively well. Second, the best-performing CorPipe system has been tuned for CorefUD over years, while LLM approaches had only a few months of testing. Third, and perhaps most importantly, we are still at the beginning of learning how to best provide LLMs with coreference-annotated data and how to elicit coreference reasoning, questions that clearly require further exploration.

\section{Conclusions and Future Work}
\label{sec:conclusions}

The paper summarizes the fourth edition of the shared task on multilingual coreference resolution, organized in 2025. Besides relatively conservative (though important too) updates with respect to the previous editions, such as improved quality of the data integrated in CorefUD and the increased number of languages, the major innovation in this edition was the support for LLM-based solutions. With only a few exceptions, LLM-based solutions did not outperform CorPipeEnsemble, the best Unconstrained system (from the same author as the winning submissions in the previous editions). However, we believe that the lower performance of the LLM solutions should be rather attributed to our currently limited knowledge of how coreference is handled internally in LLMs, and that studying how to deal with coreference in LLMs may -- in a longer-term perspective --  result in rethinking how we should represent coreference in NLP in general.

\section*{Acknowledgements}

This work has been supported
by Charles University Research Centre program No. 24/SSH/009,
Ministry of Education, Youth, and Sports of the Czech Republic, Project No.
LM2023062 LINDAT/CLARIAH-CZ and
CZ.02.01.01/00/23\_020/0008518,
and the Grant 20-16819X (LUSyD) of the Czech Science Foundation (GAČR).
We thank all the participants of the shared task for participating and for
providing brief descriptions of their systems.
We thank Kirill Milintsevich for the initial conversion of French-ANCOR into CorefUD, and Ian Porada for his assistance with the conversion of Korean-ECMT.
We also thank anonymous reviewers for their useful remarks.

\newpage %

\bibliography{anthology,custom}

\begin{thebibliography}{53}
\expandafter\ifx\csname natexlab\endcsname\relax\def\natexlab#1{#1}\fi

\bibitem[{Bagga and Baldwin(1998)}]{Bcubed-score}
Amit Bagga and Breck Baldwin. 1998.
\newblock {Algorithms for Scoring Coreference Chains}.
\newblock In \emph{Proceedings of The First International Conference on
  Language Resources and Evaluation Workshop on Linguistics Coreference}, pages
  563--566.

\bibitem[{Bamman et~al.(2020)Bamman, Lewke, and Mansoor}]{Bamman20Litbank}
David Bamman, Olivia Lewke, and Anya Mansoor. 2020.
\newblock \href {https://aclanthology.org/2020.lrec-1.6/} {{An Annotated
  Dataset of Coreference in {E}nglish Literature}}.
\newblock In \emph{Proceedings of the Twelfth Language Resources and Evaluation
  Conference}, pages 44--54, Marseille, France. European Language Resources
  Association.

\bibitem[{Bourgonje and Stede(2020)}]{potsdamCC-2020}
Peter Bourgonje and Manfred Stede. 2020.
\newblock \href {https://aclanthology.org/2020.lrec-1.133/} {{The {P}otsdam
  Commentary Corpus 2.2: Extending Annotations for Shallow Discourse Parsing}}.
\newblock In \emph{Proceedings of the Twelfth Language Resources and Evaluation
  Conference}, pages 1061--1066, Marseille, France. European Language Resources
  Association.

\bibitem[{Conneau et~al.(2020)Conneau, Khandelwal, Goyal, Chaudhary, Wenzek,
  Guzm{\'a}n, Grave, Ott, Zettlemoyer, and
  Stoyanov}]{conneau-etal-2020-unsupervised}
Alexis Conneau, Kartikay Khandelwal, Naman Goyal, Vishrav Chaudhary, Guillaume
  Wenzek, Francisco Guzm{\'a}n, Edouard Grave, Myle Ott, Luke Zettlemoyer, and
  Veselin Stoyanov. 2020.
\newblock \href {https://doi.org/10.18653/v1/2020.acl-main.747} {Unsupervised
  Cross-lingual Representation Learning at Scale}.
\newblock In \emph{Proceedings of the 58th Annual Meeting of the Association
  for Computational Linguistics}, pages 8440--8451, Online. Association for
  Computational Linguistics.

\bibitem[{Denis and Baldridge(2009)}]{CoNLL-MELA-score}
Pascal Denis and Jason Baldridge. 2009.
\newblock Global joint models for coreference resolution and named entity
  classification.
\newblock \emph{Procesamiento del lenguaje natural}, 42.

\bibitem[{Dobrovolskii(2021)}]{dobrovolskii-2021-word}
Vladimir Dobrovolskii. 2021.
\newblock \href {https://doi.org/10.18653/v1/2021.emnlp-main.605} {Word-Level
  Coreference Resolution}.
\newblock In \emph{Proceedings of the 2021 Conference on Empirical Methods in
  Natural Language Processing}, pages 7670--7675, Online and Punta Cana,
  Dominican Republic. Association for Computational Linguistics.

\bibitem[{D{'}Oosterlinck et~al.(2023)D{'}Oosterlinck, Bitew, Papineau, Potts,
  Demeester, and Develder}]{doosterlinck-etal-2023-caw}
Karel D{'}Oosterlinck, Semere~Kiros Bitew, Brandon Papineau, Christopher Potts,
  Thomas Demeester, and Chris Develder. 2023.
\newblock \href {https://doi.org/10.18653/v1/2023.crac-main.2} {{CAW}-coref:
  Conjunction-Aware Word-level Coreference Resolution}.
\newblock In \emph{Proceedings of the Sixth Workshop on Computational Models of
  Reference, Anaphora and Coreference (CRAC 2023)}, pages 8--14, Singapore.
  Association for Computational Linguistics.

\bibitem[{Emami et~al.(2019)Emami, Trichelair, Trischler, Suleman, Schulz, and
  Cheung}]{emami.etal19}
Ali Emami, Paul Trichelair, Adam Trischler, Kaheer Suleman, Hannes Schulz, and
  Jackie Chi~Kit Cheung. 2019.
\newblock \href {https://doi.org/10.18653/v1/P19-1386} {The {{KnowRef
  Coreference Corpus}}: {{Removing Gender}} and {{Number Cues}} for {{Difficult
  Pronominal Anaphora Resolution}}}.
\newblock In \emph{Proceedings of the 57th {{Annual Meeting}} of the
  {{Association}} for {{Computational Linguistics}}}, pages 3952--3961,
  Florence, Italy. Association for Computational Linguistics.

\bibitem[{Gan et~al.(2024)Gan, Poesio, and Yu}]{gan_assessing_2024}
Yujian Gan, Massimo Poesio, and Juntao Yu. 2024.
\newblock \href {https://aclanthology.org/2024.lrec-main.145/} {Assessing the
  {Capabilities} of {Large} {Language} {Models} in {Coreference}: {An}
  {Evaluation}}.
\newblock In \emph{Proceedings of the 2024 {Joint} {International} {Conference}
  on {Computational} {Linguistics}, {Language} {Resources} and {Evaluation}
  ({LREC}-{COLING} 2024)}, pages 1645--1665, Torino, Italia. ELRA and ICCL.

\bibitem[{Gevers et~al.(2025)Gevers, De~Marez, De~Bruyne, and
  Daelemans}]{gevers.etal25}
Ine Gevers, Victor De~Marez, Luna De~Bruyne, and Walter Daelemans. 2025.
\newblock \href {https://doi.org/10.18653/v1/2025.conll-1.5} {{{WinoWhat}}: {{A
  Parallel Corpus}} of {{Paraphrased WinoGrande Sentences}} with {{Common Sense
  Categorization}}}.
\newblock In \emph{Proceedings of the 29th {{Conference}} on {{Computational
  Natural Language Learning}}}, pages 68--80, Vienna, Austria. Association for
  Computational Linguistics.

\bibitem[{Haji{\v{c}} et~al.(2020)Haji{\v{c}}, Bej{\v{c}}ek,
  Hlav{\'{a}}{\v{c}}ov{\'{a}}, Mikulov{\'{a}}, Straka, {\v{S}}těp{\'{a}}nek,
  and {\v{S}}t{\v{e}}p{\'{a}}nkov{\'{a}}}]{pdtconsolidated}
Jan Haji{\v{c}}, Eduard Bej{\v{c}}ek, Jaroslava Hlav{\'{a}}{\v{c}}ov{\'{a}},
  Marie Mikulov{\'{a}}, Milan Straka, Jan {\v{S}}těp{\'{a}}nek, and Barbora
  {\v{S}}t{\v{e}}p{\'{a}}nkov{\'{a}}. 2020.
\newblock \href {https://www.aclweb.org/anthology/2020.lrec-1.641.pdf} {{Prague
  Dependency Treebank - Consolidated 1.0}}.
\newblock In \emph{Proceedings of the 12th International Conference on Language
  Resources and Evaluation (LREC 2020)}, pages 5208--5218, Marseille, France.
  European Language Resources Association.

\bibitem[{Haug and J{\o}hndal(2008)}]{Haug2008CreatingAP}
Dag Trygve~Truslew Haug and Marius~L. J{\o}hndal. 2008.
\newblock \href {https://api.semanticscholar.org/CorpusID:204978005} {{Creating
  a Parallel Treebank of the Old Indo-European Bible Translations}}.
\newblock In \emph{Proceedings of the second workshop on language technology
  for cultural heritage data (LaTeCH 2008)}.

\bibitem[{Hicke and Mimno(2024)}]{hicke_lions_2024}
Rebecca Hicke and David Mimno. 2024.
\newblock \href {https://aclanthology.org/2024.latechclfl-1.27/} {[{Lions}: 1]
  and [{Tigers}: 2] and [{Bears}: 3], {Oh} {My}! {Literary} {Coreference}
  {Annotation} with {LLMs}}.
\newblock In \emph{Proceedings of the 8th {Joint} {SIGHUM} {Workshop} on
  {Computational} {Linguistics} for {Cultural} {Heritage}, {Social} {Sciences},
  {Humanities} and {Literature} ({LaTeCH}-{CLfL} 2024)}, pages 270--277, St.
  Julians, Malta. Association for Computational Linguistics.

\bibitem[{Landragin(2021)}]{democrat}
Fr{\'e}d{\'e}ric Landragin. 2021.
\newblock \href {https://hal.archives-ouvertes.fr/hal-03474748} {{Le corpus
  Democrat et son exploitation. Pr{\'e}sentation}}.
\newblock \emph{{Langages}}, 224:11--24.

\bibitem[{Lapshinova-Koltunski et~al.(2018)Lapshinova-Koltunski, Hardmeier, and
  Krielke}]{ParCorFullScheme}
Ekaterina Lapshinova-Koltunski, Christian Hardmeier, and Pauline Krielke. 2018.
\newblock \href {https://aclanthology.org/L18-1065/} {{ParCorFull: a Parallel
  Corpus Annotated with Full Coreference}}.
\newblock In \emph{Proceedings of the Eleventh International Conference on
  Language Resources and Evaluation (LREC 2018)}, Miyazaki, Japan. European
  Language Resources Association.

\bibitem[{Le and Ritter(2023)}]{le_are_2023}
Nghia~T. Le and Alan Ritter. 2023.
\newblock \href {https://doi.org/10.48550/arXiv.2305.14489} {Are {Large}
  {Language} {Models} {Robust} {Coreference} {Resolvers}?}
\newblock ArXiv:2305.14489 [cs].

\bibitem[{Lee et~al.(2017)Lee, He, Lewis, and Zettlemoyer}]{lee-etal-2017-end}
Kenton Lee, Luheng He, Mike Lewis, and Luke Zettlemoyer. 2017.
\newblock \href {https://doi.org/10.18653/v1/D17-1018} {End-to-end Neural
  Coreference Resolution}.
\newblock In \emph{Proceedings of the 2017 Conference on Empirical Methods in
  Natural Language Processing}, pages 188--197, Copenhagen, Denmark.
  Association for Computational Linguistics.

\bibitem[{Levesque et~al.(2012)Levesque, Davis, and
  Morgenstern}]{levesque.etal12}
Hector~J. Levesque, Ernest Davis, and Leora Morgenstern. 2012.
\newblock The {{Winograd}} Schema Challenge.
\newblock In \emph{Proceedings of the {{Thirteenth International Conference}}
  on {{Principles}} of {{Knowledge Representation}} and {{Reasoning}}},
  {{KR}}'12, pages 552--561, Rome, Italy. AAAI Press.

\bibitem[{Liu et~al.(2024)Liu, Bauer, D{'}Oosterlinck, Potts, and
  Manning}]{liu-etal-2024-mscaw}
Houjun Liu, John Bauer, Karel D{'}Oosterlinck, Christopher Potts, and
  Christopher~D. Manning. 2024.
\newblock \href {https://doi.org/10.18653/v1/2024.crac-1.4} {{MSCAW}-coref:
  Multilingual, Singleton and Conjunction-Aware Word-Level Coreference
  Resolution}.
\newblock In \emph{Proceedings of the Seventh Workshop on Computational Models
  of Reference, Anaphora and Coreference}, pages 33--40, Miami. Association for
  Computational Linguistics.

\bibitem[{Luo(2005)}]{CEAF-score}
Xiaoqiang Luo. 2005.
\newblock \href {https://doi.org/10.3115/1220575.1220579} {On Coreference
  Resolution Performance Metrics}.
\newblock In \emph{Proceedings of the Conference on Human Language Technology
  and Empirical Methods in Natural Language Processing}, {HLT} 2005, pages
  25--32. Association for Computational Linguistics.

\bibitem[{M{\ae}hlum et~al.(2022)M{\ae}hlum, Haug, J{\o}rgensen, K{\aa}sen,
  N{\o}klestad, R{\o}nningstad, Solberg, Velldal, and
  {\O}vrelid}]{maehlum2022narc}
Petter M{\ae}hlum, Dag Haug, Tollef J{\o}rgensen, Andre K{\aa}sen, Anders
  N{\o}klestad, Egil R{\o}nningstad, Per~Erik Solberg, Erik Velldal, and Lilja
  {\O}vrelid. 2022.
\newblock \href {https://aclanthology.org/2022.crac-1.6/} {{NARC}--{N}orwegian
  Anaphora Resolution Corpus}.
\newblock In \emph{Proceedings of the Fifth Workshop on Computational Models of
  Reference, Anaphora and Coreference}, pages 48--60, Gyeongju, Korea.
  Association for Computational Linguistics.

\bibitem[{Moosavi and Strube(2016)}]{LEA-score}
Nafise~Sadat Moosavi and Michael Strube. 2016.
\newblock \href {https://doi.org/10.18653/v1/P16-1060} {{Which Coreference
  Evaluation Metric Do You Trust? A Proposal for a Link-based Entity Aware
  Metric}}.
\newblock In \emph{Proceedings of the 54th Annual Meeting of the Association
  for Computational Linguistics (Volume 1: Long Papers)}, pages 632--642,
  Berlin, Germany. Association for Computational Linguistics.

\bibitem[{Mujadia et~al.(2016)Mujadia, Gupta, and
  Sharma}]{mujadia-etal-2016-coreference}
Vandan Mujadia, Palash Gupta, and Dipti~Misra Sharma. 2016.
\newblock \href {https://aclanthology.org/L16-1025/} {Coreference Annotation
  Scheme and Relation Types for {H}indi}.
\newblock In \emph{Proceedings of the Tenth International Conference on
  Language Resources and Evaluation ({LREC}'16)}, pages 161--168,
  Portoro{\v{z}}, Slovenia. European Language Resources Association (ELRA).

\bibitem[{Muzerelle et~al.(2014)Muzerelle, Lefeuvre, Schang, Antoine,
  Pelletier, Maurel, Eshkol, and Villaneau}]{muzerelle-etal-2014-ancor}
Judith Muzerelle, Ana{\"i}s Lefeuvre, Emmanuel Schang, Jean-Yves Antoine,
  Aurore Pelletier, Denis Maurel, Iris Eshkol, and Jeanne Villaneau. 2014.
\newblock \href {https://aclanthology.org/L14-1169/} {{ANCOR}{\_}{C}entre, a
  large free spoken {F}rench coreference corpus: description of the resource
  and reliability measures}.
\newblock In \emph{Proceedings of the Ninth International Conference on
  Language Resources and Evaluation ({LREC}'14)}, pages 843--847, Reykjavik,
  Iceland. European Language Resources Association (ELRA).

\bibitem[{Nam et~al.(2020)Nam, Lee, Kim, Han, Kim, Yoon, Kim, and
  Choi}]{nam-etal-2020-effective}
Sangha Nam, Minho Lee, Donghwan Kim, Kijong Han, Kuntae Kim, Sooji Yoon,
  Eun-kyung Kim, and Key-Sun Choi. 2020.
\newblock \href {https://aclanthology.org/2020.lrec-1.27/} {Effective
  Crowdsourcing of Multiple Tasks for Comprehensive Knowledge Extraction}.
\newblock In \emph{Proceedings of the Twelfth Language Resources and Evaluation
  Conference}, pages 212--219, Marseille, France. European Language Resources
  Association.

\bibitem[{Nedoluzhko et~al.(2016)Nedoluzhko, Nov{\'a}k, Cinkov{\'a},
  Mikulov{\'a}, and M{\'\i}rovsk{\'y}}]{PCEDT2016coreference}
Anna Nedoluzhko, Michal Nov{\'a}k, Silvie Cinkov{\'a}, Marie Mikulov{\'a}, and
  Ji{\v{r}}{\'\i} M{\'\i}rovsk{\'y}. 2016.
\newblock \href {https://www.aclweb.org/anthology/L16-1026} {Coreference in
  {P}rague {C}zech-{E}nglish {D}ependency {T}reebank}.
\newblock In \emph{Proceedings of the Tenth International Conference on
  Language Resources and Evaluation ({LREC} 2016)}, pages 169--176,
  Portoro{\v{z}}, Slovenia. European Language Resources Association.

\bibitem[{Nedoluzhko et~al.(2022)Nedoluzhko, Nov{\'a}k, Popel,
  {\v{Z}}abokrtsk{\'y}, Zeldes, and Zeman}]{corefud2022lrec}
Anna Nedoluzhko, Michal Nov{\'a}k, Martin Popel, Zden{\v{e}}k
  {\v{Z}}abokrtsk{\'y}, Amir Zeldes, and Daniel Zeman. 2022.
\newblock \href {https://aclanthology.org/2022.lrec-1.520} {{C}oref{UD} 1.0:
  Coreference Meets {U}niversal {D}ependencies}.
\newblock In \emph{Proceedings of the Thirteenth Language Resources and
  Evaluation Conference}, pages 4859--4872, Marseille, France. European
  Language Resources Association.

\bibitem[{Nov{\'a}k et~al.(2024)Nov{\'a}k, Dohnalov{\'a}, Konopik, Nedoluzhko,
  Popel, Prazak, Sido, Straka, {\v{Z}}abokrtsk{\'y}, and
  Zeman}]{oursharedtask2024}
Michal Nov{\'a}k, Barbora Dohnalov{\'a}, Miloslav Konopik, Anna Nedoluzhko,
  Martin Popel, Ondrej Prazak, Jakub Sido, Milan Straka, Zden{\v{e}}k
  {\v{Z}}abokrtsk{\'y}, and Daniel Zeman. 2024.
\newblock \href {https://doi.org/10.18653/v1/2024.crac-1.8} {Findings of the
  Third Shared Task on Multilingual Coreference Resolution}.
\newblock In \emph{Proceedings of the Seventh Workshop on Computational Models
  of Reference, Anaphora and Coreference}, pages 78--96, Miami. Association for
  Computational Linguistics.

\bibitem[{Nov{\'a}k et~al.(2025)Nov{\'a}k, Popel, Zeman, {\v Z}abokrtsk{\'y},
  Nedoluzhko, Acar, Bamman, Bourgonje, Cinkov{\'a}, Eckhoff, Cebiro{\u
  g}lu~Eryi{\u g}it, Haji{\v c}, Hardmeier, Haug, J{\o}rgensen, K{\aa}sen,
  Krielke, Landragin, Lapshinova-Koltunski, M{\ae}hlum, Mart{\'{\i}},
  Mikulov{\'a}, Milintsevich, Mujadia, Muzerelle, Nam, N{\o}klestad,
  Ogrodniczuk, {\O}vrelid, Pamay~Arslan, Porada, Recasens, Solberg, Stede,
  Straka, Swanson, Toldova, Vad{\'a}sz, Velldal, Vincze, Zeldes, and {\v
  Z}itkus}]{corefud1.3}
Michal Nov{\'a}k, Martin Popel, Daniel Zeman, Zden{\v e}k {\v Z}abokrtsk{\'y},
  Anna Nedoluzhko, Kutay Acar, David Bamman, Peter Bourgonje, Silvie
  Cinkov{\'a}, Hanne Eckhoff, G{\"u}l{\c s}en Cebiro{\u g}lu~Eryi{\u g}it, Jan
  Haji{\v c}, Christian Hardmeier, Dag Haug, Tollef J{\o}rgensen, Andre
  K{\aa}sen, Pauline Krielke, Fr{\'e}d{\'e}ric Landragin, Ekaterina
  Lapshinova-Koltunski, Petter M{\ae}hlum, M.~Ant{\`o}nia Mart{\'{\i}}, Marie
  Mikulov{\'a}, Kirill Milintsevich, Vandan Mujadia, Judith Muzerelle, Sangha
  Nam, Anders N{\o}klestad, Maciej Ogrodniczuk, Lilja {\O}vrelid, Tu{\u g}ba
  Pamay~Arslan, Ian Porada, Marta Recasens, Per~Erik Solberg, Manfred Stede,
  Milan Straka, Daniel Swanson, Svetlana Toldova, No{\'e}mi Vad{\'a}sz, Erik
  Velldal, Veronika Vincze, Amir Zeldes, and Voldemaras {\v Z}itkus. 2025.
\newblock \href {http://hdl.handle.net/11234/1-5896} {Coreference in Universal
  Dependencies 1.3 ({CorefUD} 1.3)}.
\newblock {LINDAT}/{CLARIAH}-{CZ} digital library at the Institute of Formal
  and Applied Linguistics ({{\'U}FAL}), Faculty of Mathematics and Physics,
  Charles University.

\bibitem[{Ogrodniczuk et~al.(2013)Ogrodniczuk, Glowińska, Kopeć, Savary, and
  Zawisławska}]{PCC2013}
Maciej Ogrodniczuk, Katarzyna Glowińska, Mateusz Kopeć, Agata Savary, and
  Magdalena Zawisławska. 2013.
\newblock \href {https://doi.org/10.1007/978-3-319-43808-5\_17} {{Polish
  Coreference Corpus}}.
\newblock In \emph{Human Language Technology. Challenges for Computer Science
  and Linguistics --- 6th Language and Technology Conference ({LTC} 2013),
  Revised Selected Papers}, volume 9561 of \emph{Lecture Notes in Computer
  Science}, pages 215--226. Springer.

\bibitem[{Ogrodniczuk et~al.(2015)Ogrodniczuk, Głowińska, Kopeć, Savary, and
  Zawisławska}]{bookOgrodniczuk}
Maciej Ogrodniczuk, Katarzyna Głowińska, Mateusz Kopeć, Agata Savary, and
  Magdalena Zawisławska. 2015.
\newblock \href {http://www.degruyter.com/view/product/428667}
  {\emph{{Coreference in {P}olish: Annotation, Resolution and Evaluation}}}.
\newblock Walter De Gruyter.

\bibitem[{Palmer et~al.(2009)Palmer, Bhatt, Narasimhan, Rambow, Sharma, and
  Xia}]{palmer2009hindi}
Martha Palmer, Rajesh Bhatt, Bhuvana Narasimhan, Owen Rambow, Dipti~Misra
  Sharma, and Fei Xia. 2009.
\newblock Hindi syntax: Annotating dependency, lexical predicate-argument
  structure, and phrase structure.
\newblock In \emph{The 7th International Conference on Natural Language
  Processing}, pages 14--17.

\bibitem[{Pamay and Eryi{\u{g}}it(2018)}]{pamay2018coref}
Tu{\u{g}}ba Pamay and G{\"u}l{\c{s}}en Eryi{\u{g}}it. 2018.
\newblock \href {https://doi.org/10.1109/INISTA.2018.8466293} {{Turkish
  Coreference Resolution}}.
\newblock In \emph{2018 Innovations in Intelligent Systems and Applications
  (INISTA)}, pages 1--7.

\bibitem[{Pradhan et~al.(2014)Pradhan, Luo, Recasens, Hovy, Ng, and
  Strube}]{pradhan-etal-2014-scoring}
Sameer Pradhan, Xiaoqiang Luo, Marta Recasens, Eduard Hovy, Vincent Ng, and
  Michael Strube. 2014.
\newblock \href {https://doi.org/10.3115/v1/P14-2006} {Scoring Coreference
  Partitions of Predicted Mentions: A Reference Implementation}.
\newblock In \emph{Proceedings of the 52nd Annual Meeting of the Association
  for Computational Linguistics (Volume 2: Short Papers)}, pages 30--35,
  Baltimore, Maryland. Association for Computational Linguistics.

\bibitem[{Pra{\v{z}}{\'a}k et~al.(2021)Pra{\v{z}}{\'a}k, Konop{\'i}k, and
  Sido}]{prazak-etal-2021-multilingual}
Ond{\v{r}}ej Pra{\v{z}}{\'a}k, Miloslav Konop{\'i}k, and Jakub Sido. 2021.
\newblock \href {https://aclanthology.org/2021.ranlp-1.125/} {Multilingual
  Coreference Resolution with Harmonized Annotations}.
\newblock In \emph{Proceedings of the International Conference on Recent
  Advances in Natural Language Processing (RANLP 2021)}, pages 1119--1123, Held
  Online. INCOMA Ltd.

\bibitem[{Recasens et~al.(2010)Recasens, Hovy, and
  Mart{\'\i}}]{recasensNearIdentity2010}
Marta Recasens, Eduard Hovy, and M.~Ant{\`o}nia Mart{\'\i}. 2010.
\newblock \href
  {http://www.lrec-conf.org/proceedings/lrec2010/pdf/160_Paper.pdf} {{A
  Typology of Near-Identity Relations for Coreference ({NIDENT})}}.
\newblock In \emph{Proceedings of the Seventh International Conference on
  Language Resources and Evaluation ({LREC} 2010)}, Valletta, Malta. European
  Language Resources Association.

\bibitem[{Recasens and Hovy(2011)}]{BLANC-score}
Marta Recasens and Eduard~H. Hovy. 2011.
\newblock \href {https://doi.org/10.1017/S135132491000029X} {{{BLANC}:
  Implementing the Rand index for coreference evaluation}}.
\newblock \emph{Natural Language Engineering}, 17(4):485--510.

\bibitem[{Recasens and Mart\'{\i}(2010)}]{ancora-co}
Marta Recasens and M.~Ant\`{o}nia Mart\'{\i}. 2010.
\newblock \href {https://doi.org/10.1007/s10579-009-9108-x} {{AnCora-CO:
  Coreferentially Annotated Corpora for Spanish and Catalan}}.
\newblock \emph{Language Resources and Evaluation}, 44(4):315–345.

\bibitem[{Sakaguchi et~al.(2021)Sakaguchi, Bras, Bhagavatula, and
  Choi}]{sakaguchi.etal21}
Keisuke Sakaguchi, Ronan~Le Bras, Chandra Bhagavatula, and Yejin Choi. 2021.
\newblock \href {https://doi.org/10.1145/3474381} {{{WinoGrande}}: An
  Adversarial Winograd Schema Challenge at Scale}.
\newblock \emph{Commun. ACM}, 64(9):99--106.

\bibitem[{Saputa et~al.(2024)Saputa, Peljak-Łapińska, and
  Ogrodniczuk}]{saputa_polish_2024}
Karol Saputa, Angelika Peljak-Łapińska, and Maciej Ogrodniczuk. 2024.
\newblock \href {https://doi.org/10.18653/v1/2024.crac-1.3} {Polish
  {Coreference} {Corpus} as an {LLM} {Testbed}: {Evaluating} {Coreference}
  {Resolution} within {Instruction}-{Following} {Language} {Models} by
  {Instruction}–{Answer} {Alignment}}.
\newblock In \emph{Proceedings of {The} {Seventh} {Workshop} on {Computational}
  {Models} of {Reference}, {Anaphora} and {Coreference}}, pages 23--32, Miami.
  Association for Computational Linguistics.

\bibitem[{Straka(2024)}]{straka-2024-corpipe}
Milan Straka. 2024.
\newblock \href {https://doi.org/10.18653/v1/2024.crac-1.9} {{C}or{P}ipe at
  {CRAC} 2024: Predicting Zero Mentions from Raw Text}.
\newblock In \emph{Proceedings of the Seventh Workshop on Computational Models
  of Reference, Anaphora and Coreference}, pages 97--106, Miami. Association
  for Computational Linguistics.

\bibitem[{Swanson et~al.(2024)Swanson, Bussert, and
  Tyers}]{swanson-etal-2024-towards}
Daniel~G. Swanson, Bryce~D. Bussert, and Francis Tyers. 2024.
\newblock \href {https://aclanthology.org/2024.lt4hala-1.5/} {Towards
  Named-Entity and Coreference Annotation of the {H}ebrew {B}ible}.
\newblock In \emph{Proceedings of the Third Workshop on Language Technologies
  for Historical and Ancient Languages (LT4HALA) @ LREC-COLING-2024}, pages
  36--40, Torino, Italia. ELRA and ICCL.

\bibitem[{Taul{\'e} et~al.(2008)Taul{\'e}, Mart{\'\i}, and Recasens}]{ancora}
Mariona Taul{\'e}, M.~Ant{\`o}nia Mart{\'\i}, and Marta Recasens. 2008.
\newblock \href
  {http://www.lrec-conf.org/proceedings/lrec2008/pdf/35_paper.pdf}
  {{{A}n{C}ora: Multilevel Annotated Corpora for {C}atalan and {S}panish}}.
\newblock In \emph{Proceedings of the Sixth International Conference on
  Language Resources and Evaluation ({LREC} 2008)}, Marrakech, Morocco.
  European Language Resources Association.

\bibitem[{Toldova et~al.(2014)Toldova, Roytberg, Ladygina, Vasilyeva,
  Azerkovich, Kurzukov, Sim, Gorshkov, Ivanova, Nedoluzhko, and
  Grishina}]{RuCorDialog}
Svetlana Toldova, Anna Roytberg, Alina Ladygina, Maria Vasilyeva, Ilya
  Azerkovich, Matvei Kurzukov, G.~Sim, D.V. Gorshkov, A.~Ivanova, Anna
  Nedoluzhko, and Yulia Grishina. 2014.
\newblock {Evaluating Anaphora and Coreference Resolution for Russian}.
\newblock In \emph{Komp'juternaja lingvistika i intellektual'nye tehnologii. Po
  materialam ezhegodnoj Mezhdunarodnoj konferencii Dialog}, pages 681--695.

\bibitem[{Vad{\'a}sz(2022)}]{korkor_coling}
No{\'e}mi Vad{\'a}sz. 2022.
\newblock \href {https://aclanthology.org/2022.crac-1.5/} {{Building a Manually
  Annotated {H}ungarian Coreference Corpus: Workflow and Tools}}.
\newblock In \emph{Proceedings of the Fifth Workshop on Computational Models of
  Reference, Anaphora and Coreference}, pages 38--47, Gyeongju, Korea.
  Association for Computational Linguistics.

\bibitem[{Vad\'{a}sz(2023)}]{vadasz2023resolving}
No\'{e}mi Vad\'{a}sz. 2023.
\newblock \href {https://doi.org/10.1007/978-3-031-40498-6_5} {{Resolving
  Hungarian Anaphora with ChatGPT}}.
\newblock In \emph{Text, Speech, and Dialogue: 26th International Conference,
  TSD 2023, Pilsen, Czech Republic, September 4–6, 2023, Proceedings}, page
  45–57, Berlin, Heidelberg. Springer-Verlag.

\bibitem[{Vilain et~al.(1995)Vilain, Burger, Aberdeen, Connolly, and
  Hirschman}]{MUC-score}
Marc Vilain, John Burger, John Aberdeen, Dennis Connolly, and Lynette
  Hirschman. 1995.
\newblock \href {https://aclanthology.org/M95-1005} {{A Model-Theoretic
  Coreference Scoring Scheme}}.
\newblock In \emph{Sixth Message Understanding Conference ({MUC}-6):
  Proceedings of a Conference Held in {C}olumbia, {M}aryland, November 6-8,
  1995}.

\bibitem[{Vincze et~al.(2018)Vincze, Heged{\H{u}}s, Sliz-Nagy, and
  Farkas}]{szegedkoref2018}
Veronika Vincze, Kl{\'a}ra Heged{\H{u}}s, Alex Sliz-Nagy, and Rich{\'a}rd
  Farkas. 2018.
\newblock \href {https://www.aclweb.org/anthology/L18-1061} {{{S}zeged{K}oref:
  A {H}ungarian Coreference Corpus}}.
\newblock In \emph{Proceedings of the Eleventh International Conference on
  Language Resources and Evaluation ({LREC} 2018)}, Miyazaki, Japan. European
  Language Resources Association.

\bibitem[{Yu et~al.(2023)Yu, Nov{\'a}k, Aloraini, Moosavi, Paun, Pradhan, and
  Poesio}]{ua-scorer-2.0}
Juntao Yu, Michal Nov{\'a}k, Abdulrahman Aloraini, Nafise~Sadat Moosavi, Silviu
  Paun, Sameer Pradhan, and Massimo Poesio. 2023.
\newblock \href {https://aclanthology.org/2023.iwcs-1.19} {{The Universal
  Anaphora Scorer 2.0}}.
\newblock In \emph{Proceedings of the 15th International Conference on
  Computational Semantics}, pages 183--194, Nancy, France. Association for
  Computational Linguistics.

\bibitem[{{\v{Z}}abokrtsk{\'y} et~al.(2023){\v{Z}}abokrtsk{\'y}, Konop{\'\i}k,
  Nedoluzhko, Nov{\'a}k, Ogrodniczuk, Popel, Pra{\v{z}}{\'a}k, Sido, and
  Zeman}]{oursharedtask2023}
Zden{\v{e}}k {\v{Z}}abokrtsk{\'y}, Miloslav Konop{\'\i}k, Anna Nedoluzhko,
  Michal Nov{\'a}k, Maciej Ogrodniczuk, Martin Popel, Ond{\v{r}}ej
  Pra{\v{z}}{\'a}k, Jakub Sido, and Daniel Zeman. 2023.
\newblock \href {https://doi.org/10.18653/v1/2023.crac-sharedtask.1} {{Findings
  of the Second Shared Task on Multilingual Coreference Resolution}}.
\newblock In \emph{Proceedings of the CRAC 2023 Shared Task on Multilingual
  Coreference Resolution}, pages 1--18, Singapore. Association for
  Computational Linguistics.

\bibitem[{{\v{Z}}abokrtsk{\'y} et~al.(2022){\v{Z}}abokrtsk{\'y}, Konop{\'\i}k,
  Nedoluzhko, Nov{\'a}k, Ogrodniczuk, Popel, Pra{\v{z}}{\'a}k, Sido, Zeman, and
  Zhu}]{oursharedtask2022}
Zden{\v{e}}k {\v{Z}}abokrtsk{\'y}, Miloslav Konop{\'\i}k, Anna Nedoluzhko,
  Michal Nov{\'a}k, Maciej Ogrodniczuk, Martin Popel, Ond{\v{r}}ej
  Pra{\v{z}}{\'a}k, Jakub Sido, Daniel Zeman, and Yilun Zhu. 2022.
\newblock \href {https://aclanthology.org/2022.crac-mcr.1/} {{Findings of the
  Shared Task on Multilingual Coreference Resolution}}.
\newblock In \emph{Proceedings of the CRAC 2022 Shared Task on Multilingual
  Coreference Resolution}, pages 1--17, Gyeongju, Korea. Association for
  Computational Linguistics.

\bibitem[{Zeldes(2017)}]{GUMdocumentation}
Amir Zeldes. 2017.
\newblock \href {https://doi.org/10.1007/s10579-016-9343-x} {{The GUM Corpus:
  Creating Multilayer Resources in the Classroom}}.
\newblock \emph{Language Resources and Evaluation}, 51(3):581--612.

\bibitem[{{\v Z}itkus and Butkien\.{e}(2018)}]{LithuanianScheme}
Voldemaras {\v Z}itkus and Rita Butkien\.{e}. 2018.
\newblock \href {https://doi.org/10.1109/SNAMS.2018.8554892} {{Coreference
  Annotation Scheme and Corpus for {Lithuanian} Language}}.
\newblock In \emph{Fifth International Conference on Social Networks Analysis,
  Management and Security, {SNAMS} 2018, Valencia, Spain, October 15-18, 2018},
  pages 243--250. {IEEE}.

\end{thebibliography}
\bibliographystyle{acl_natbib}

\appendix
\clearpage
\onecolumn

\section{CorefUD 1.3 Details}
\label{sec:data-references}
\begin{tabular}{llll}
  Ancient Greek  & PROIEL      & \grcproiel{}     & \cite{Haug2008CreatingAP} \\
  Ancient Hebrew  & PTNK       & \hboptnk{}       & \cite{swanson-etal-2024-towards} \\
  Catalan    & AnCora          & \caancora{}      & \cite{ancora,ancora-co} \\
  Czech      & PCEDT           & \cspcedt{}       & \cite{PCEDT2016coreference} \\
  Czech      & PDT             & \cspdt{}         & \cite{pdtconsolidated} \\
  English    & GUM             & \engum{}         & \cite{GUMdocumentation} \\
  English    & LitBank         & \enlitbank{}     & \cite{Bamman20Litbank} \\
  English    & ParCorFull      & \enparcorfull{}  & \cite{ParCorFullScheme} \\
  French     & ANCOR           & \francor{}       & \cite{muzerelle-etal-2014-ancor} \\
  French     & Democrat        & \frdemocrat{}    & \cite{democrat} \\
  German     & ParCorFull      & \deparcorfull{}  & \cite{ParCorFullScheme} \\
  German     & PotsdamCC       & \depotsdamcc{}   & \cite{potsdamCC-2020} \\
  Hindi      & HDTB            & \hihdtb{}        & \cite{mujadia-etal-2016-coreference} \\
  Hungarian  & KorKor          & \hukorkor{}      & \cite{korkor_coling} \\
  Hungarian  & SzegedKoref     & \huszegedkoref{} & \cite{szegedkoref2018} \\
  Korean     & ECMT            & \koecmt{}        & \cite{nam-etal-2020-effective} \\
  Lithuanian & LCC             & \ltlcc{}         & \cite{LithuanianScheme} \\
  Norwegian  & Bokmål NARC     & \nobokmaalnarc{} & \cite{maehlum2022narc} \\
  Norwegian  & Nynorsk NARC    & \nonynorsknarc{} & \cite{maehlum2022narc} \\
  Old Church Slavonic & PROIEL & \cuproiel        & \cite{Haug2008CreatingAP} \\
  Polish     & PCC             & \plpcc{}         & \cite{PCC2013,bookOgrodniczuk} \\
  Russian    & RuCor           & \rurucor{}       & \cite{RuCorDialog} \\
  Spanish    & AnCora          & \esancora{}      & \cite{ancora,ancora-co} \\
  Turkish    & ITCC            & \tritcc          & \cite{pamay2018coref} \\
\end{tabular}

\section{CoNLL results by head UPOS}
\label{sec:stats-upos}

\begin{table}[H]\centering
\begin{tabular}{@{}l r@{~~~}r@{~~~}r@{~~~}r@{~~~}r@{~~~}r@{~~~}r@{~~~}r@{}}\toprule
\bf system         & \bf NOUN  & \bf PRON  & \bf PROPN & \bf DET   & \bf ADJ   & \bf VERB  & \bf ADV   & \bf NUM  \\\midrule
CorPipeEnsemble    & \bf 71.78 & \bf 71.67 & \bf 78.11 &     52.58 &     47.92 & \bf 37.36 & \bf 32.03 &     37.40 \\
CorPipeBestDev     &     71.07 &     71.13 &     77.69 &     49.22 & \bf 48.35 &     36.62 &     27.62 &     38.22 \\
CorPipeSingle      &     70.96 &     70.47 &     77.28 & \bf 53.01 &     44.69 &     35.45 &     31.96 & \bf 38.76 \\
Stanza             &     62.55 &     64.24 &     70.94 &     41.78 &     32.77 &     21.73 &     21.89 &     29.58 \\
LLM-GLaRef-CRAC25  &     58.81 &     61.23 &     64.30 &     41.83 &     29.26 &     23.08 &     20.90 &     34.52 \\
LLM-NUST-FewShot   &     58.01 &     59.21 &     69.88 &     32.79 &     34.39 &     14.39 &     20.59 &     26.36 \\
GLaRef-Propp       &     56.44 &     57.99 &     63.20 &     36.10 &     28.43 &     17.88 &     20.26 &     21.56 \\
LLM-PUXCRAC2025    &     54.71 &     56.22 &     64.51 &     36.55 &     27.53 &     15.36 &     17.86 &     25.76 \\
LLM-UWB            &     57.19 &     55.95 &     64.72 &     36.83 &     29.57 &     22.30 &     23.53 &     26.25 \\
\baselinegz        &     50.74 &     58.46 &     57.21 &     37.24 &     25.85 &     14.15 &     18.15 &     23.11 \\
\baseline          &     48.44 &     52.03 &     54.96 &     36.75 &     24.04 &     13.44 &     16.98 &     22.81 \\
\bottomrule\end{tabular}

\caption{CoNLL F$_1$ score (head-match) evaluated only on entities with heads of a given
UPOS. In both the gold and prediction files we deleted some entities before
running the evaluation. We kept only entities with at least one mention with
a given head UPOS (universal part of speech tag). For the purpose of this
analysis,
 if the head node had deprel=flat children,
 their UPOS tags were considered as well,
 so for example in ``Mr./NOUN Brown/PROPN''
 both NOUN and PROPN were taken as head UPOS,
 so the entity with this mention will be reported in both columns NOUN and PROPN.
Otherwise, the CoNLL F$_1$ scores are the same as in the primary metric,
 i.e.\ an unweighted average over all datasets, head-match, without singletons.
Note that when distinguishing entities into events and nominal entities,
 the VERB column can be considered as an approximation of the performance on events.
One of the limitations of this approach is that copula is not treated as head in the Universal Dependencies,
 so, e.g., phrase \textit{She is nice} is not considered for the VERB column,
 but for the ADJ column (head of the phrase is \textit{nice}).
}
\label{tab:upos-entity}
\end{table}

\begin{table}[H]\centering
\begin{tabular}{@{}l r@{~~~}r@{~~~}r@{~~~}r@{~~~}r@{~~~}r@{~~~}r@{~~~}r@{}}\toprule
\bf system         & \bf NOUN  & \bf PRON  & \bf PROPN & \bf DET   & \bf ADJ   & \bf VERB  & \bf ADV   & \bf NUM  \\\midrule
CorPipeEnsemble    & \bf 63.91 & \bf 61.69 & \bf 64.74 & \bf 53.28 & \bf 51.12 & \bf 50.58 & \bf 50.81 & \bf 50.46 \\
CorPipeBestDev     &     62.42 &     60.85 &     63.57 &     52.51 &     49.91 &     48.72 &     49.33 &     49.00 \\
CorPipeSingle      &     62.91 &     60.69 &     64.05 &     52.66 &     49.98 &     49.66 &     49.92 &     49.72 \\
Stanza             &     54.67 &     54.66 &     56.77 &     44.31 &     42.51 &     41.37 &     42.31 &     41.78 \\
LLM-GLaRef-CRAC25  &     50.80 &     51.80 &     52.12 &     41.98 &     39.11 &     38.75 &     39.08 &     38.81 \\
LLM-NUST-FewShot   &     52.16 &     52.84 &     54.26 &     42.09 &     40.05 &     39.47 &     40.28 &     39.96 \\
GLaRef-Propp       &     47.57 &     48.85 &     49.46 &     36.41 &     33.83 &     33.37 &     34.09 &     33.58 \\
LLM-PUXCRAC2025    &     47.37 &     46.07 &     49.09 &     34.88 &     33.11 &     31.91 &     32.71 &     32.48 \\
LLM-UWB            &     51.82 &     47.99 &     53.14 &     40.23 &     37.45 &     36.91 &     37.44 &     36.99 \\
\baselinegz        &     42.44 &     49.49 &     45.96 &     33.76 &     31.16 &     30.43 &     31.05 &     30.61 \\
\baseline          &     40.99 &     42.45 &     44.50 &     31.94 &     29.42 &     28.58 &     29.17 &     28.80 \\
\bottomrule\end{tabular}
\caption{CoNLL F$_1$ score (head-match) evaluated only on mentions with heads of a given UPOS.
In both the gold and prediction files we deleted some mentions before running the evaluation.
We kept only mentions with a given head UPOS (again considering also deprel=flat children).
These results may be a bit
misleading because e.g.\ the PRON column does not consider all pronominal
coreference, but only pronoun-to-pronoun coreference. An entity with one
pronoun and one noun mention is excluded from this table (because it becomes
a singleton after deleting noun or pronoun mentions and singletons are
excluded from the evaluation in this table).
}
\label{tab:upos-mention}
\end{table}

\section{Statistics of the submitted systems on concatenation of all test sets}
\label{sec:stats-concat}
The systems are sorted alphabetically in tables in this section.

\begin{table}[H]\centering
\begin{tabular}{@{}l rrrr rrrrr@{}}\toprule
                    & \MC{4}{entities}                  & \MC{5}{distribution of lengths}      \\\cmidrule(lr){2-5}\cmidrule(l){6-10}
system              &   total & per 1k & \MC{2}{length} &     1 &     2 &     3 &     4 &   5+ \\\cmidrule(lr){4-5}
                    &   count &  words &    max &  avg. &  [\%] &  [\%] &  [\%] &  [\%] & [\%] \\\midrule
gold                &  39,576 &    108 &    509 &   2.1 &  67.4 &  17.3 &   5.9 &   2.8 &   6.6 \\    
\baseline           &  10,591 &     29 &    347 &   4.2 &   0.0 &  55.8 &  17.6 &   7.8 &  18.9 \\    
\baselinegz         &  10,977 &     30 &    354 &   4.2 &   0.0 &  55.5 &  17.6 &   7.8 &  19.2 \\    
CorPipeBestDev      &  40,392 &    111 &    248 &   2.1 &  66.6 &  17.7 &   6.2 &   2.8 &   6.6 \\    
CorPipeEnsemble     &  40,615 &    111 &    461 &   2.0 &  66.5 &  17.8 &   6.3 &   2.9 &   6.5 \\    
CorPipeSingle       &  40,377 &    111 &    362 &   2.1 &  66.6 &  17.7 &   6.2 &   3.0 &   6.6 \\    
GLaRef-Propp        &  40,481 &    111 &    563 &   1.9 &  75.0 &  12.4 &   4.6 &   2.3 &   5.7 \\    
LLM-GLaRef-CRAC25   &  39,664 &    109 &    280 &   1.9 &  70.6 &  15.1 &   5.6 &   2.7 &   6.0 \\    
LLM-NUST-FewShot    &  35,703 &     98 &    393 &   2.0 &  71.1 &  13.5 &   5.5 &   2.8 &   7.1 \\    
LLM-PUXCRAC2025     &  19,896 &     55 &    545 &   2.9 &  44.3 &  29.4 &  10.1 &   4.8 &  11.5 \\    
LLM-UWB             &  35,542 &     97 &    317 &   1.9 &  70.0 &  15.6 &   5.6 &   2.8 &   6.0 \\    
Stanza              &  38,464 &    105 &    523 &   2.0 &  67.8 &  17.4 &   5.9 &   2.8 &   6.2 \\    
\bottomrule\end{tabular}
\caption{Statistics on coreference entities.
The total number of entities and the average number of entities per 1000 tokens in the running text.
The maximum and average entity ``length'',
 i.e., the number of mentions in the entity.
Distribution of entity lengths (singletons have length = 1).
The two baselines and LLM-PUXCRAC2025 heavily undergenerate
 (i.e.\ predict less entities than in the gold data)
 and the baselines also predict on average longer entities (i.e.\ with more mentions) than in the gold data.
The remaining systems have the statistics similar to the gold data,
 (although the CorPipe* systems and GLaRef-Propp slightly overgenerate,
 while LLM-NUST-FewShot and LLM-UWB undergenerate).
}
\label{tab:stats-entities}
\end{table}

\begin{table}[H]\centering
\begin{tabular}{@{}l rrrr rrrrrr@{}}\toprule
                    & \MC{4}{non-singleton mentions}                  & \MC{6}{distribution of lengths}              \\\cmidrule(lr){2-5}\cmidrule(l){6-11}
system              &   total & per 1k & \MC{2}{length} &     0 &     1 &     2 &     3 &     4 &   5+ \\\cmidrule(lr){4-5}
                    &   count &  words &    max &  avg. &  [\%] &  [\%] &  [\%] &  [\%] &  [\%] & [\%] \\\midrule
gold                &  55,333 &    152 &    100 &   2.5 &   9.8 &  50.1 &  19.1 &   7.0 &   3.3 &  10.8 \\    
\baseline           &  44,110 &    121 &     27 &   1.9 &  10.0 &  54.9 &  18.8 &   6.3 &   2.6 &   7.3 \\    
\baselinegz         &  45,989 &    126 &     27 &   1.9 &  11.4 &  54.2 &  18.5 &   6.2 &   2.6 &   7.1 \\    
CorPipeBestDev      &  56,020 &    154 &    149 &   2.4 &   9.6 &  51.0 &  19.1 &   6.9 &   3.1 &  10.3 \\    
CorPipeEnsemble     &  55,668 &    153 &    149 &   2.4 &   9.6 &  51.0 &  19.0 &   6.9 &   3.1 &  10.2 \\    
CorPipeSingle       &  56,026 &    154 &    140 &   2.5 &   9.6 &  50.9 &  19.1 &   6.9 &   3.1 &  10.4 \\    
GLaRef-Propp        &  48,362 &    133 &     51 &   1.9 &   9.9 &  55.3 &  19.2 &   6.4 &   2.6 &   6.6 \\    
LLM-GLaRef-CRAC25   &  49,311 &    135 &     96 &   2.3 &  10.7 &  52.1 &  18.6 &   6.4 &   3.0 &   9.2 \\    
LLM-NUST-FewShot    &  47,681 &    131 &    104 &   2.0 &   6.9 &  58.0 &  19.1 &   6.2 &   2.6 &   7.2 \\    
LLM-PUXCRAC2025     &  48,593 &    133 &     27 &   1.8 &   8.4 &  57.8 &  18.4 &   5.9 &   2.5 &   6.9 \\    
LLM-UWB             &  42,852 &    117 &     58 &   1.8 &   1.2 &  80.6 &   8.3 &   2.9 &   1.4 &   5.6 \\    
Stanza              &  50,811 &    139 &    100 &   2.3 &   9.3 &  52.8 &  18.9 &   6.6 &   2.9 &   9.6 \\    
\bottomrule\end{tabular}
\caption{Statistics on non-singleton mentions.
The total number of mentions and the average number of
mentions per 1000 words of running text. The maximum and average mention length, i.e., the number of nonempty nodes (words) in the mention. Distribution of mention lengths (zeros have length = 0).
Only the CorPipe* systems generate a similar number of non-singleton mentions as in the gold data,
 all other systems generate less mentions.
The average length of mentions predicted by LLM-UWB is notably lower than in the gold data
 because LLM-UWB predicted single-word mentions only
 in most datasets.
All other systems have the distribution of mention lengths similar to the gold data,
although no system predicts long mentions (4 and 5+ words) more frequently than in the gold data,
 (but CorPipe* systems are near to the gold distribution).
}
\label{tab:stats-mentions-nonsingleton}
\end{table}

\begin{table}[H]\centering
\begin{tabular}{@{}l rrrr rrrrrr@{}}\toprule
                    & \MC{4}{singleton mentions}                  & \MC{6}{distribution of lengths}              \\\cmidrule(lr){2-5}\cmidrule(l){6-11}
system              &   total & per 1k & \MC{2}{length} &     0 &     1 &     2 &     3 &     4 &   5+ \\\cmidrule(lr){4-5}
                    &   count &  words &    max &  avg. &  [\%] &  [\%] &  [\%] &  [\%] &  [\%] & [\%] \\\midrule
gold                &  26,661 &     73 &     81 &   3.0 &   0.7 &  39.4 &  24.0 &  12.2 &   6.3 &  17.3 \\    
\baseline           &       0 &      0 &      0 &   0.0 &   0.0 &   0.0 &   0.0 &   0.0 &   0.0 &   0.0 \\    
\baselinegz         &       0 &      0 &      0 &   0.0 &   0.0 &   0.0 &   0.0 &   0.0 &   0.0 &   0.0 \\    
CorPipeBestDev      &  26,919 &     74 &    112 &   3.1 &   0.7 &  38.1 &  24.8 &  12.7 &   6.4 &  17.3 \\    
CorPipeEnsemble     &  27,014 &     74 &    112 &   3.0 &   0.7 &  38.5 &  25.0 &  12.5 &   6.2 &  17.0 \\    
CorPipeSingle       &  26,885 &     74 &     85 &   3.1 &   0.7 &  38.5 &  24.9 &  12.6 &   6.3 &  17.1 \\    
GLaRef-Propp        &  30,343 &     83 &     33 &   2.3 &   2.4 &  40.4 &  27.7 &  13.0 &   6.1 &  10.5 \\    
LLM-GLaRef-CRAC25   &  28,021 &     77 &     80 &   2.9 &   0.9 &  40.5 &  25.1 &  12.2 &   5.8 &  15.5 \\    
LLM-NUST-FewShot    &  25,379 &     70 &     63 &   2.8 &   0.2 &  41.8 &  24.9 &  12.0 &   5.9 &  15.3 \\    
LLM-PUXCRAC2025     &   8,807 &     24 &     17 &   2.0 &   0.4 &  52.5 &  23.8 &  11.4 &   4.1 &   7.8 \\    
LLM-UWB             &  24,889 &     68 &     86 &   1.7 &   0.0 &  78.2 &  10.0 &   4.3 &   2.1 &   5.4 \\    
Stanza              &  26,060 &     71 &    100 &   2.9 &   1.4 &  40.2 &  24.5 &  11.8 &   6.1 &  16.0 \\    
\bottomrule\end{tabular}
\caption{Statistics on singleton mentions.
See the caption of Table~\ref{tab:stats-mentions-nonsingleton} for details.
The two baseline systems do not attempt to predict singletons at all.
LLM-PUXCRAC2025 heavily undergenerates singletons.
GLaRef-Propp overgenerates singletons (including zeros),
 but note that singletons are not annotated in all the (gold) datasets.
}
\label{tab:stats-mentions-singleton}
\end{table}

\begin{table}[H]\centering
\resizebox{\textwidth}{!}{
\begin{tabular}{@{}l @{\!}r@{~}r@{~}r @{~}r@{~}r@{~}r@{~}r@{~}r@{~}r@{~}r@{~}r@{~}r@{~}r@{}}\toprule
                    & \MC{3}{mention type [\%]}    & \MC{9}{distribution of head UPOS [\%]}      \\\cmidrule(lr){2-4}\cmidrule(l){5-14}
system              & w/empty & w/gap & non-tree
                                             &  NOUN &  PRON & PROPN &   DET &   ADJ &  VERB &   ADV &   NUM & \_~ & other \\\midrule
gold                &  11.0 &   0.7 &   1.4 &  38.6 &  31.5 &  17.7 &   4.2 &   1.3 &   1.9 &   1.4 &   0.5 &   2.1 &   0.8 \\    
\baseline           &  10.5 &   0.0 &   1.4 &  35.4 &  26.9 &  18.7 &   4.8 &   1.1 &   0.9 &   1.2 &   0.4 &  10.0 &   0.6 \\    
\baselinegz         &  12.0 &   0.0 &   1.5 &  35.1 &  34.7 &  18.5 &   4.7 &   1.1 &   0.9 &   1.5 &   0.4 &   2.5 &   0.8 \\    
CorPipeBestDev      &  10.6 &   0.0 &   1.9 &  39.0 &  23.7 &  17.6 &   4.3 &   1.2 &   1.8 &   1.4 &   0.5 &   9.6 &   0.8 \\    
CorPipeEnsemble     &  10.6 &   0.0 &   1.8 &  39.0 &  23.8 &  17.7 &   4.3 &   1.2 &   1.7 &   1.4 &   0.5 &   9.6 &   0.8 \\    
CorPipeSingle       &  10.5 &   0.0 &   1.9 &  39.1 &  23.7 &  17.6 &   4.3 &   1.2 &   1.7 &   1.4 &   0.5 &   9.6 &   0.8 \\    
GLaRef-Propp        &   9.9 &   0.0 &   1.4 &  35.5 &  26.9 &  18.4 &   4.7 &   1.1 &   0.8 &   1.4 &   0.4 &   9.9 &   0.9 \\    
LLM-GLaRef-CRAC25   &  11.4 &   0.0 &   1.8 &  37.5 &  24.7 &  17.0 &   4.7 &   1.3 &   1.4 &   1.4 &   0.5 &  10.7 &   1.0 \\    
LLM-NUST-FewShot    &   7.1 &   0.0 &   1.3 &  39.4 &  25.9 &  18.6 &   3.5 &   1.2 &   1.5 &   1.5 &   0.5 &   6.9 &   1.1 \\    
LLM-PUXCRAC2025     &   8.9 &   0.0 &   1.4 &  37.2 &  25.7 &  18.7 &   4.1 &   1.3 &   2.0 &   1.3 &   0.5 &   8.4 &   0.8 \\    
LLM-UWB             &   1.2 &   0.0 &   0.8 &  42.9 &  24.6 &  20.9 &   4.8 &   1.3 &   1.1 &   1.7 &   0.5 &   1.2 &   1.0 \\    
Stanza              &  10.0 &   0.0 &   1.4 &  39.0 &  24.0 &  18.8 &   4.1 &   1.1 &   1.1 &   1.4 &   0.4 &   9.3 &   0.8 \\    
\bottomrule\end{tabular}
}
\caption{Detailed statistics on non-singleton mentions.
The left part of the table shows the percentage of:
 mentions with at least one empty node (w/empty);
 mentions with at least one gap, i.e.\ discontinuous mentions (w/gap);
 and non-treelet mentions, i.e.\ mentions not forming a connected subgraph (catena) in the dependency tree (non-tree).
Note that these three types of mentions may be overlapping.
We can see that none of the systems attempts to predict discontinuous mentions.
LLM-UWB has a notably lower percentage (0.8\%) of non-treelet mention spans,
 but this is simply explained by its higher percentage (80\%) of single-word mentions.
The right part of the table shows the distribution of mentions
  based on the universal part-of-speech tag (UPOS) of the head word.
Note that this distribution has to be interpreted with the total number of non-singleton mentions predicted (as reported in Table~\ref{tab:stats-mentions-nonsingleton}) in mind.
For example, 34.7\% of non-singleton mentions predicted by \baselinegz{} are pronominal (head=PRON),
  while there are only 31.5\% of pronominal non-singleton mentions in the gold data.
However, \baselinegz{} predicts actually less pronominal non-singleton mentions 
 ($45{,}989 \times 34.7\% \approx 15{,}958$) than in the gold data ($55{,}333 \times 31.5\% \approx 17{,}430$).
Note that the same word may be assigned a different UPOS tag in the predicted and gold data
(in case of empty nodes or if the gold data includes manual annotation).
The empty UPOS tag (\_) is present only in the empty nodes
and none of the systems attempts to predict the actual UPOS tag of empty nodes
(they all keep the empty tag from the baseline predictor of empty nodes,
although about 78\% of the empty nodes in the gold devset are pronouns).
}
\label{tab:stats-details}
\end{table}

\clearpage
\section{Evolution of CodaLab Submissions}
\label{sec:codalab-evol}
\begin{figure*}[h]
    \centering
    \includegraphics[width=0.7\textwidth]{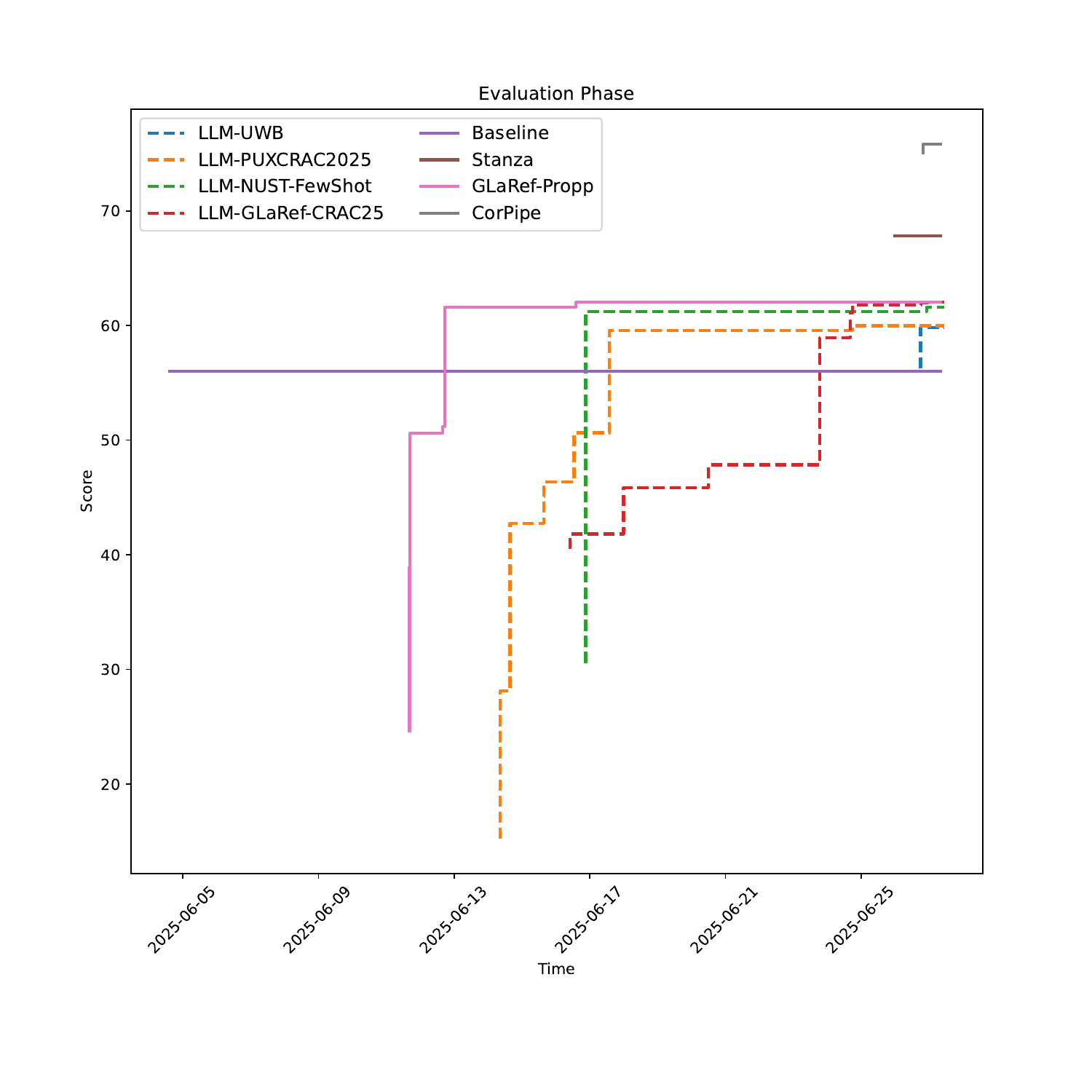}
    \caption{Evolution of CodaLab Submissions in the %
    evaluation phase. The submissions to the LLM and Unconstrained track are shown by using the dashed and solid lines, respectively. For clarity, all submissions of the ÚFAL CorPipe team are represented by the scores of CorPipeEnsemble.}
    \label{fig:codalab-evol}
\end{figure*}

\end{document}